\documentclass[11pt]{article}

\usepackage[a4paper,margin=1in]{geometry}
\usepackage{amsmath,amssymb,amsfonts,amsthm}
\usepackage{graphicx}
\usepackage{hyperref}
\usepackage{bm}
\usepackage{enumerate}
\usepackage{mathrsfs}

\usepackage{amsthm}

\usepackage{float}

\usepackage{placeins}

\usepackage{tikz}
\usepackage{pgfplots}
\usetikzlibrary{arrows,shapes,positioning,calc}
\usepackage{subcaption}

\hypersetup{
  colorlinks=true,
  linkcolor=blue,
  citecolor=blue,
  urlcolor=blue
}

\usepackage{amsmath,amssymb,amsthm}

\numberwithin{equation}{section}

\theoremstyle{plain}
\newtheorem{theorem}{Theorem}[section]

\theoremstyle{definition}

\newtheorem{definition}{Definition}[section]

\numberwithin{figure}{section}
\numberwithin{table}{section}

\title{Why Inference in Large Models Becomes Decomposable After Training}

\author{
Jidong Jin\thanks{School of Artificial Intelligence,
Capital University of Economics and Business, Beijing 100070, China. \\ Email: jjd@cueb.edu.cn, jidongjin@Gmail.com}
}

\date{2026-01}

\begin{document}

\maketitle

\begin{abstract}

Current large-scale AI models typically perform inference by operating on the full parameter matrix as a single matrix operator.
As model size continues to grow, this paradigm leads to inference costs and system complexity that scale in an increasingly unsustainable manner.
The core issue does not lie in the expressive capacity of the model itself, but rather in the fact that the post-training inference system has long been treated as a structurally uniform operator,
while the internal structural differences that emerge during learning have not been explicitly identified.

Under over-parameterized conditions, gradient updates exhibit strong locality and selectivity in parameter space.
As a result, the parameter matrix obtained after training usually contains a large number of parameters that have not received sustained support from the training samples.
These parameters fail to accumulate stable gradient updates,
and their statistical behavior remains close to the initialization distribution even after training.
Consequently, the post-training inference system is not structurally uniform;
instead, it naturally exhibits a decomposable substructure.

Based on this observation,
the present work takes the parameter initialization distribution as a statistical null hypothesis in the post-training stage
and constructs a posterior decision mechanism to distinguish dependency relations that have been significantly confirmed by the learning process
from parameter components that have not formed structural effects.
By applying statistical annealing to the latter,
the original full-matrix parameter system can be transformed into a representation with pronounced block structural characteristics,
revealing relatively independent sub-operator structures.

The proposed framework operates solely on parameter matrices
and does not rely on specific model implementation details.
Without altering the model's functionality or its input--output interface,
it enables explicit identification and stabilization of the structure formed after training,
thereby providing an engineering-operational pathway for structural reorganization,
parallel execution, and complexity control of inference systems.

\end{abstract}

\part{Structural Annealing and System Reorganization}

\section{Introduction}

In recent years, large-scale artificial neural networks, 
particularly deep learning models and foundation-scale systems, 
have demonstrated remarkable performance in perception, reasoning, and generation tasks
\cite{lecun2015deep,brown2020language,wei2022chain,
bommasani2021foundation,srivastava2023beyond}.

At both the algorithmic and engineering levels,
such systems typically take high-dimensional parameter matrices as their core computational objects.
After training is completed,
the parameter organization formed during the training phase
is usually carried directly into long-term inference and deployment
\cite{vaswani2017attention,cheng2018model,
narayanan2021efficient,dao2022flashattention}.
As model size continues to grow,
this paradigm of operating the model as a single \emph{global matrix operator}
gradually reveals its unsustainability
in terms of inference cost, system complexity, and energy consumption
\cite{kaplan2020scaling,strubell2019energy,
rae2022scaling,chowdhery2022palm}.

Meanwhile,
a large body of research has shown that,
under over-parameterized conditions,
gradient updates in neural networks exhibit strong locality and selectivity in parameter space,
and systematically favor solutions with particular structural characteristics
\cite{zhang2017rethinking,neyshabur2017exploring,
gunasekar2018implicit,chizat2020lazy,yang2023feature}.
Model analyses and circuit studies have also revealed that
after training,
stable functional organizations and internal substructures
emerge within neural models
\cite{olah2018building,olah2020zoom,elhage2021circuits}.

However,
existing approaches either focus on structural design during the model construction stage
(such as neural architecture search)
\cite{zoph2016neural,elsken2019neural,wen2016structured},
or perform numerical pruning and compression after training
\cite{cheng2018model,han2015learning,frankle2019lottery,
hoefler2021sparsity,liu2019rethinking,liu2023sparse}.
None of these approaches explicitly model the
\emph{posterior structural organization formed by the learning process itself}
as an independent system-level layer.

The focus of this work is therefore not how models should be designed during training,
but rather what structural form the inference system has already taken
once training is completed.
We propose a posterior structural reorganization framework
based entirely on parameter matrices.
Without altering model semantics or the input--output interface,
the framework explicitly identifies and stabilizes dependency relations
that have been confirmed during learning,
transforming an inference system that originally operates as a single global operator
into a composite system composed of multiple relatively independent sub-operators.

We refer to this process as the
\emph{structural stabilization of the inference system after learning in large AI models}.
The goal of this work is to establish a complete, verifiable,
and engineering-operable theoretical framework
to characterize and realize the posterior structural independence of inference systems,
thereby providing a structural pathway
for complexity control and system-level scalability
in ultra-large-scale models.

\section{Feasibility of Post-Training Reorganization}

The structural reorganization framework proposed in this work
is grounded in a series of phenomena that have been repeatedly observed empirically.
These phenomena are not incidental results of specific architectures or tasks,
but rather stable statistical characteristics
that consistently appear after the training of large neural models.

\paragraph{Non-uniform utilization of parameters.}

A number of studies have shown that,
although parameter matrices appear dense in form,
only a subset of parameters continuously receive effective gradient support during training,
while the update magnitude of many parameters
remains statistically close to the level of initialization noise
\cite{frankle2019lottery,
hoefler2021sparsity,liu2019rethinking,liu2023sparse,michel2019sixteen}.
This phenomenon is not merely introduced by explicit sparsification methods,
but arises naturally from gradient optimization
under structured data distributions.

\paragraph{Input-dependent activation bias.}

It has also been observed that
different categories of inputs
tend to consistently activate different attention heads,
feed-forward channels,
or parameter subspaces
\cite{voita2019analyzing,vig2019analyzing,
geva2021transformer,dalvi2020analyzing}.
These observations indicate that,
after training,
stable functional usage patterns of internal subspaces
have already emerged within the model.

\paragraph{Structural differentiation under homogeneous architectures.}

Notably,
the above non-uniformity does not depend on explicit modular architectural design.
Even in highly homogeneous Transformer architectures,
parameter contributions and subspace activity
still exhibit clear differentiation after training
\cite{dalvi2020analyzing}.
This suggests that structural differences arise
from the joint effects of optimization dynamics and data statistics,
rather than from manually imposed structural priors.

Taken together,
these observations indicate that
the inference system after training
is not a uniform whole,
but rather consists of several weakly coupled substructures.
These substructures receive sustained statistical support during training
and remain stable during inference.
The structural reorganization proposed in this work
is precisely a posterior identification and formalization
of this objectively existing structural state.

Based on the above abstraction,
the structural reorganization problem studied in this work
uniformly applies to artificial neural systems
whose core modules consist of collections of parameter matrices,
or to their structurally equivalent representations.

\section{Inference-System Annealing}

The objects considered in this work include various artificial neural networks and their variants,
such as feed-forward neural networks (FNN),
as well as large-scale deep learning systems constructed upon them
\cite{lecun2015deep,goodfellow2016deep}.
Although these models differ in implementation and engineering details,
they share a common structural feature:
their core computational modules
can be uniformly expressed
as an operator family
composed of parameter matrices.

Within this abstraction,
artificial neurons may be viewed as state variables,
while connections between neurons
are represented by the elements (parameters) of matrices.
The parameter values associated with edges
characterize the strength and direction
of the influence that upstream neurons exert
on the states of downstream neurons
\cite{goodfellow2016deep}.

\subsection{Initial Global Statistical Test (Neyman Significance Test)}

All parameters in the trained parameter matrices
are treated as a sample set of test objects
\[
\mathcal{W}=\{w_{ij}\}.
\]
The statistical decision of structural annealing
can naturally be formulated
as a hypothesis test in the Neyman framework:
for each parameter $w_{ij}$,
we determine whether it significantly deviates
from the null hypothesis of
``no structural effect.''

During the parameter initialization stage,
parameters are typically independently sampled
from a prescribed initialization distribution
$f_0(w)$
(for example, a symmetric zero-mean distribution).
This distribution characterizes the random level
that parameters would exhibit
\emph{in the absence of any learning process},
and therefore serves as the theoretical null hypothesis
for structural determination.

\medskip
\noindent
Null hypothesis ($H_0$):  
Parameter $w_{ij}$ still follows the initialization distribution $f_0(w)$,
reflecting only random initialization or noise perturbation,
and does not correspond to a structural dependency confirmed by the learning process.

\noindent
Alternative hypothesis ($H_1$):  
Parameter $w_{ij}$ significantly deviates from the initialization distribution,
indicating that the learning process has formed
a stable structural effect along this direction.

\medskip
Given a significance level $x\%\,(0<x\ll100)$,
the rejection region is determined by $f_0(w)$,
for example by the two-sided tail interval:
\[
|w_{ij}|\ge c_x,
\qquad
\mathbb{P}_{f_0}(|w|\ge c_x)=x\%.
\]

\medskip
\noindent
According to the Neyman tail $x\%$ criterion,
if $w_{ij}$ falls within the rejection region,
$H_0$ is rejected,
and the corresponding edge $j\to i$
is regarded as a structurally significant effective edge.
Otherwise,
$H_0$ is accepted,
and the edge is considered not to have been significantly confirmed by the learning process,
and may therefore be annealed at the structural level.

This test serves to distinguish
\emph{which edges have been significantly altered during learning and which have not};
however,
it does not distinguish
the specific source or mechanism
responsible for the parameter deviation.

\subsection{Random-Walk Hypothesis Test (Equiprobable Distribution Test)}

After learning is completed,
each parameter $w_{ij}$ in the parameter matrix $W=[w_{ij}]_{n\times n}$
represents the strength
of the influence exerted by node $j$ on node $i$.
Such influence may arise either from structural dependency
or merely from directionless random perturbation.

To characterize the latter situation,
we introduce the random-walk noise hypothesis:
when node $j$ has not formed a stable structural preference,
its influence on different nodes $i$
can be regarded as an equiprobable random allocation.

Under this assumption,
the parameter matrix is normalized as follows:
\[
\tilde w_{ij}
=
\frac{|w_{ij}|}{\sum_{i=1}^n |w_{ij}|},
\qquad
\sum_{i=1}^n \tilde w_{ij}=1.
\]

In the ideal random-walk scenario,
we should have
\[
\tilde w_{ij}\approx \frac{1}{n},
\]
meaning that different directions are statistically indistinguishable.

\medskip
\noindent
To transform the above random-walk hypothesis
into an operational statistical decision rule,
it is first necessary to determine
the admissible bandwidth $\delta$
of equiprobable random perturbations.
The procedure is as follows.

All normalized parameters
\[
\tilde{\mathcal W}
=
\{\tilde w_{ij}\}
\]
are treated as a global sample set.
For a given candidate bandwidth $\delta>0$,
consider the subset
\[
\tilde{\mathcal W}(\delta)
=
\left\{
\tilde w_{ij}\in\tilde{\mathcal W}
\;\middle|\;
\tilde w_{ij}\in
\left[
\frac{1}{n}-\delta,\;
\frac{1}{n}+\delta
\right]
\right\}.
\]

Under the random-walk null hypothesis,
$\tilde{\mathcal W}(\delta)$
should follow an equiprobable distribution
centered at $\frac{1}{n}$.
Therefore,
within the Neyman--Pearson framework,
a goodness-of-fit test may be applied to
$\tilde{\mathcal W}(\delta)$,
for example using the Pearson $\chi^2$ test
or other equivalent distribution tests.

The testing procedure adopts
a progressively expanding interval strategy.
Starting from a small initial bandwidth $\delta_0$,
if the random-walk null hypothesis cannot be rejected
at the given significance level $\alpha$,
the interval width is gradually increased,
letting
\[
\delta_{k+1}=\delta_k+\tau,
\]
and the test is repeated.
When the null hypothesis is rejected for the first time,
the bandwidth $\delta_k$
from the previous round in which the null hypothesis was still accepted
is taken as the maximal admissible bandwidth
of equiprobable random-walk noise,
denoted by $\delta$.

\medskip
\noindent
Under this bandwidth $\delta$,
the normalized parameters can be classified into three categories:
\[
\begin{array}{ll}
(1) & \tilde w_{ij}>\frac{1}{n}+\delta \quad\text{(structural preference)},\\[4pt]
(2) & \tilde w_{ij}\in\left[\frac{1}{n}-\delta,\frac{1}{n}+\delta\right]
\quad\text{(random-walk noise)},\\[6pt]
(3) & \tilde w_{ij}<\frac{1}{n}-\delta
\quad\text{(systematic suppression)}.
\end{array}
\]

The purpose of this test
is not to compare the magnitude of parameters,
but to distinguish the \emph{source and nature}
of parameter changes:
whether they arise from structural preference,
directionless random perturbation,
or directions that have been systematically suppressed
by the learning process.

\subsection{Summary of the Annealing Rules}

In summary, the structural annealing method for artificial neural networks proposed in this work
follows a unified and clear underlying logic:
\emph{parameters are taken as the objects of annealing,
statistical significance serves as the decision criterion,
and the objective is to elevate the effective dependency relations
implicitly formed during learning
from the parameter level
to stable structural entities.}

Structural annealing can be constructed directly
on the basis of the parameter initialization distribution
through a Neyman-style significance test.
This test determines whether a parameter significantly deviates
from the random level corresponding to
``no structural effect,''
thereby naturally partitioning parameters
into effective edges confirmed by the learning process
and ineffective edges that can be annealed at the structural level.

However, the mere deviation of parameters from the initialization distribution
is insufficient to characterize whether such deviations possess directional structure.
To address this issue,
the present work further introduces the random-walk noise hypothesis,
normalizes parameter changes,
and tests whether they still behave as
equiprobable and directionless random perturbations.
The purpose of this test
is not to compare parameter magnitudes,
but rather to distinguish three fundamentally different situations:
structural preferences formed by the learning process,
random perturbations that are statistically indistinguishable,
and dependency directions that have been systematically suppressed during learning.
Consequently, structural annealing is no longer equivalent
to simple numerical pruning,
but becomes a structural extraction process
based on statistical decision criteria.

It should be emphasized that
the above two types of tests are logically complementary rather than interchangeable.
The former answers whether a parameter has been significantly altered by the learning process,
while the latter determines whether such alteration constitutes a structural dependency.
Only edges that pass both tests
are regarded as valid structural edges
to be retained in the subsequent stage of structural reorganization.

From a methodological perspective,
the annealing strategy proposed in this work
differs both from heuristic pruning based on parameter magnitude
and from sparsification methods that introduce regularization during training.
Instead, it constitutes a post-training structural reorganization mechanism.
This mechanism does not alter the model's input--output interface
and introduces no additional structural assumptions.
Rather, it relies strictly on the statistical distinctions
revealed by the learning process itself,
thereby providing a stable and interpretable structural foundation
for subsequent system reorganization,
modular execution,
and structural evolution.

\section{Inference-System Reorganization}

\subsection{Graph Structure of Large Parameter Matrices}

From the historical development of artificial neural networks,
traditional neural network models typically contain feedback connections.
In such systems,
the parameter matrix describing interactions among states is square,
and the $i$-th row and the $i$-th column correspond to the same neuron node.

However,
in many modern applications of neural networks,
especially in multilayer neural architectures
(such as feed-forward neural networks, FNN),
parameter matrices without feedback connections are widely used.
In these models,
the parameter matrix is usually treated as a linear operator,
where the input $X$ and the output $Y$ correspond to neuron states
belonging to different layers.
In this case,
the rows and columns of the parameter matrix correspond to different sets of nodes,
and thus no longer have a one-to-one correspondence.
Consequently,
the numbers of rows and columns of the parameter matrix may differ.

From the perspective of graph structure,
a parameter matrix with feedback connections corresponds to a directed graph,
whereas a parameter matrix without feedback connections
naturally corresponds to a bipartite graph.

We next discuss the graph structural characteristics
of these two types of artificial neural networks separately.

\subsubsection{Subgraph Structure of Neural Networks with Feedback Connections}

In a typical artificial neural network
\begin{equation}\label{EQ:ANN}
N=\langle V,\ E\subset V\times V,\ W\rangle
\end{equation}
each element $w_{ij}$ of the parameter matrix has two meanings:
it indicates both the existence of a directed edge $(i,j)$
between neurons $i$ and $j$
and the weight associated with that edge.
Regarding the existence of edges,
the following correspondence holds:
\[
(i,j) \in E \;\Leftrightarrow\; w_{ij} \neq 0 .
\]

After the parameter matrix $W$ undergoes annealing,
some parameters $w_{ij}$ are set to $0$,
and the originally dense parameter matrix becomes sparse.
As a direct consequence,
the originally fully connected network
may experience structural disconnection,
leading to several subnetworks that are internally connected
but mutually disconnected:
\begin{equation}\label{EQ:Sub-ANN}
N_r=\langle V_r,\ E_r \subset V_r \times V_r,\ W^r\rangle .
\end{equation}

In this case,
there exists a permutation $P$ such that
\begin{equation}\label{PWPT}
PWP^T  = \left( {\begin{array}{cccc}
   W^1 &        &        &        \\
       & \ddots &        &        \\
       &        & W^m    &        \\
       &        &        & \mathbf{0}
 \end{array} } \right),
\end{equation}
where $W^r\ (r=1,\dots,m)$ denotes the parameter matrix
of subnetwork (\ref{EQ:Sub-ANN}),
and the zero block in the lower-right corner
corresponds to the set of isolated nodes.
In the operator sense,
each $W^r$ forms an independent sub-operator,
and no cross-block coupling exists among them.

By choosing an appropriate permutation $P$,
the diagonal blocks in (\ref{PWPT})
can further exhibit the following triangular block structure
\cite{horn2013matrix}:
\begin{equation}\label{EQ:Frobenius}
W^r  = \left( {\begin{array}{cccccc}
   W_{11}^r &        &        &        &        &        \\
            & \ddots &        &        & \mathbf{0} &        \\
   \mathbf{0} &        & W_{pp}^r &        &        &        \\
   W_{(p+1)1}^r & \cdots & W_{(p+1)p}^r & W_{(p+1)(p+1)}^r &        &        \\
   \vdots     & \ddots & \vdots     & \vdots     & \ddots &        \\
   W_{(p+q)1}^r & \cdots & W_{(p+q)p}^r & W_{(p+q)(p+1)}^r & \cdots & W_{(p+q)(p+q)}^r
 \end{array} } \right) .
\end{equation}

Equation (\ref{EQ:Frobenius})
is the Frobenius normal form of $W^r$.
The diagonal blocks $W^{r}_{ii}\ (i=1,\dots,p+q)$
are irreducible matrices
and are called Frobenius blocks of $W^r$.
Among them,
$W^{r}_{ii}\ (i=1,\dots,p)$
are independent Frobenius blocks,
while
$W^{r}_{jj}\ (j=p+1,\dots,p+q)$
are non-independent Frobenius blocks.
The matrices $W^{r}_{ij}\ (i>p,\ j<i)$
are referred to as communication blocks
between $W^{r}_{ii}$ and $W^{r}_{jj}$.

From a graph-theoretic perspective,
equation (\ref{PWPT})
corresponds to the decomposition of the neural network
(\ref{EQ:ANN})
into weakly connected subgraphs,
while the diagonal blocks in (\ref{EQ:Frobenius})
correspond to the decomposition of each weakly connected subgraph
into strongly connected components.

\subsubsection{Subgraph Structure of Neural Networks without Feedback Connections}

In some applications of artificial neural networks,
especially in feed-forward neural networks (FNN),
the parameter matrix $W$ is directly treated as a linear operator:
\begin{equation}\label{EQ:FNN}
Y_{Output} = W X_{Input},
\end{equation}
where
$Y_{Output}= (y_1,\dots,y_s)^T$,
$X_{Input}=(x_1,\dots,x_n)^T$,
and $W \in \mathbb{R}^{s\times n}$.

In this setting,
$x_i$ and $y_p$ represent the states of two different types of neurons.
The two types of neurons have a clear directional relationship:
the state of the former serves as input to the latter,
while the latter does not serve as input to the former.
No input–output relationship exists between neurons of the same type,
that is,
neither $x_i,x_j$ nor $y_p,y_q$ form input–output relations.
Moreover,
in equation (\ref{EQ:FNN})
the numbers of rows and columns of the parameter matrix $W$
are allowed to differ.

After the parameter matrix $W$ undergoes annealing,
the originally globally connected structure
may experience structural disconnection,
leading to several subnetworks that are internally connected
but mutually disconnected.
In this case,
there exist permutations $P,Q$ such that
\begin{equation}\label{EQ:FNN-1}
PWQ^T  = \left( {\begin{array}{cccc}
   W^1 &        &        &        \\
       & \ddots &        &        \\
       &        & W^m    &        \\
       &        &        & \mathbf{0}
 \end{array} } \right).
\end{equation}

From a graph-theoretic perspective,
the parameter matrix $W$ in equation (\ref{EQ:FNN})
corresponds to a bipartite graph.
If $W$ is regarded as the parameter matrix of a neural network
without feedback connections
\[
N = \langle Y \cup X,\;E \subset Y \times X,\;W\rangle ,
\]
then each diagonal block $W^r$
in (\ref{EQ:FNN-1})
corresponds to a maximal weakly connected subnetwork of $N$:
\begin{equation}\label{EQ:2}
N_r  = \langle Y_r \cup X_r,\ E_r \subset Y_r \times X_r,\ W^r \rangle .
\end{equation}

It should be emphasized that,
although equations (\ref{PWPT}) and (\ref{EQ:FNN-1})
both exhibit block-diagonal structures,
the diagonal blocks in (\ref{PWPT}) are square matrices,
whereas in (\ref{EQ:FNN-1})
the numbers of rows and columns of each block $W^r$
may differ.

\subsection{Permutations, Projections, and Embeddings}\label{permutation}

To characterize the structural mapping between the system
before and after reorganization,
this work introduces three types of mapping operators:
permutations, projections, and embeddings.
It should be emphasized that the term “permutation”
used in this paper always refers to a bijective mapping
on a finite index set,
consistent with the standard definition in permutation group theory
\cite{Sims1971Computation, Seress2003Permutation, armstrong1988groups}.
Permutation matrices are merely one possible representation of such mappings
and are not regarded as the permutations themselves.

The permutations $P,Q$ appearing in (\ref{PWPT}) and (\ref{EQ:FNN-1})
can be interpreted as permutation matrices,
which is the most common representation of permutations
in linear algebra.
However, when permutations act on matrices or vectors,
their original meaning is the following unary bijective function
defined on row/column index positions
(note that a unary function can be regarded as an operator of a unary operation):
\[
{P < NewIndex,Index > }
\]

Since this is a finite function,
it can be represented in the following value-table form:
\begin{equation}\label{EQ:arrangement}
  P = \begin{array}{*{20}c}
   {\left( {\begin{array}{*{20}c}
   1 & {s_1 }  \\
    \vdots  &  \vdots   \\
   n & {s_n }  \\
 \end{array} } \right)} & {} & {s_i  \in \{ 1, \cdots n\} } & {} & {i \ne j \to s_i  \ne s_j }  \\
 \end{array}
\end{equation}

Let $X = (x_1 , \cdots ,x_n )^T$ and $Y = (y_1 , \cdots ,y_n )^T$.
Then the relation $Y=PX$ corresponds to
\[
\begin{array}{*{20}c}
   {Y(i) = X(s_i )} & {} & {i = 1, \cdots n}  \\
 \end{array}
\]

From an implementation perspective,
computing $Y=PX$ requires only $n$ assignment operations
(i.e., position shifts of vector components).
The permutation $P$ specifies the positional correspondence
between vector components before and after reassignment,
and therefore essentially defines a reordering of vector indices.
In fact,
even if the vector $X$ is reordered in place according to $P$
without introducing additional storage,
the algorithmic complexity remains $O(n)$.
Consequently,
the complexity of computing $PWP^T$ is $O(2n)$.
By contrast,
if the operation $Y=PX$ is implemented using a permutation matrix,
the complexity becomes $O(n^3)$,
and an additional $n\times n$ storage is required
to store the permutation matrix $P$,
whereas the actual storage cost of permutation $P$
is only $2\times n$.

Formally,
the permutation $P$ in (\ref{EQ:arrangement})
is analogous to an index in database systems.
However,
while database indices typically map
between non-primary keys and primary keys
and thus have broader meanings,
both share a common feature:
they are finite functions
that can be defined at the application level
and used as operators.

In this paper,
we define $P$ in (\ref{EQ:arrangement}) as a row permutation
(row-order transformation permutation)
acting on matrices and column vectors,
while $P^T$ is a column permutation
(column-order transformation permutation)
acting on matrices and row vectors.
$P^{-1}$ denotes the inverse permutation of $P$,
satisfying $P(P^{-1})=(P^{-1})P=I_n$.
\[
\begin{array}{*{20}c}
   {P^T  = \left( {\begin{array}{*{20}c}
   1 &  \cdots  & n  \\
   {s_1 } &  \cdots  & {s_n }  \\
 \end{array} } \right)} & {} &
 {P^{ - 1}  = \left( {\begin{array}{*{20}c}
   {s_1 } & 1  \\
    \vdots  &  \vdots   \\
   {s_n } & n  \\
 \end{array} } \right)} & {} &
 {I_n  = \left( {\begin{array}{*{20}c}
   1 & 1  \\
    \vdots  &  \vdots   \\
   n & n  \\
 \end{array} } \right)}  \\
 \end{array}
\]
Thus
\[
PWP^T  =
\left( {\begin{array}{*{20}c}
   1 & {s_1 }  \\
    \vdots  &  \vdots   \\
   n & {s_n }  \\
 \end{array} } \right)
W
\left( {\begin{array}{*{20}c}
   1 &  \cdots  & n  \\
   {s_1 } &  \cdots  & {s_n }  \\
 \end{array} } \right).
\]

corresponds to reordering the rows of $W$
from positions $s_1,\dots,s_n$ to $1,\dots,n$
while simultaneously reordering the columns
from $s_1,\dots,s_n$ to $1,\dots,n$.
This operation format is consistent
with the permutation representation
used in linear algebraic matrix operations.
However,
here permutations replace permutation matrices,
and the relation between $P,P^T$ and $W$
is not matrix multiplication
but rather the application of permutation operators
$P,P^T$
to the object $W$.

Similarly,
\[
\begin{array}{*{20}c}
   {PX} & {,} & {YP^T }.  \\
 \end{array}
\]

correspond respectively
to reordering the rows of the column vector $X$
from $s_1,\dots,s_n$ to $1,\dots,n$
and reordering the columns of the row vector $Y$
from $s_1,\dots,s_n$ to $1,\dots,n$.

In addition,
this work introduces two further operators
related to index ordering of matrices and vectors:
the projection $\pi_i$
(a restriction from higher dimension to lower dimension)
and the embedding $\gamma_i$
(an embedding from lower dimension to higher dimension).
These operators form a complementary pair
and are mutually inverse within their domains.
\[
\begin{array}{*{20}c}
   {\pi _i  = \left( {\begin{array}{*{20}c}
   1 & {s_1 }  \\
    \vdots  &  \vdots   \\
   k & {s_k }  \\
 \end{array} } \right)} &
 {} &
 {\pi _i^T  = \left( {\begin{array}{*{20}c}
   1 &  \cdots  & k  \\
   {s_1 } &  \cdots  & {s_k }  \\
 \end{array} } \right)} &
 {} &
 {\pi _i^{ - 1}  = \left( {\begin{array}{*{20}c}
   {s_1 } & 1  \\
    \vdots  &  \vdots   \\
   {s_k } & k  \\
 \end{array} } \right)}  \\
 \end{array}
\]
where
\[
\begin{array}{*{20}c}
   {\{ s_1 , \cdots ,s_k \}  \subset \{ 1, \cdots ,n\} } & {} & {i \ne j \to s_i  \ne s_j }
 \end{array}
\]

\[
\begin{array}{*{20}c}
   {B = \pi _i A,} & {B = A\pi _i^T ,} & {B = \pi _i A\pi _i^T ,} & {B = \pi _i^{ - 1} A}.
 \end{array}
\]

(1) If $A$ is a column vector or matrix,
$B=\pi_i A$ assigns the rows $1,\dots,k$ of $B$
to be rows $s_1,\dots,s_k$ of $A$.

(2) If $A$ is a row vector or matrix,
$B=A\pi_i^T$ assigns the columns $1,\dots,k$ of $B$
to be columns $s_1,\dots,s_k$ of $A$.

(3) If $A$ is a matrix,
$B=\pi_i A\pi_i^T$
extracts the submatrix of $A$
with rows and columns $s_1,\dots,s_k$.

(4) If $A$ is a column vector or matrix,
$B=\pi_i^{-1}A$
embeds rows $1,\dots,k$ of $A$
into rows $s_1,\dots,s_k$ of $B$.

In this work,
$\pi_i$ is used for the projection of input variables.
Although $\pi_i^{-1}$ is the embedding operator paired with $\pi_i$,
it is not always suitable
for embedding output variables.
Therefore,
we separately define an embedding operator $\gamma_i$
for output variables:
\[
\begin{array}{*{20}c}
   {\gamma _i  = \left( {\begin{array}{*{20}c}
   {s_1 } & 1  \\
    \vdots  &  \vdots   \\
   {s_k } & k  \\
 \end{array} } \right)} &
 , &
 {\gamma _i^T  = \left( {\begin{array}{*{20}c}
   {s_1 } &  \cdots  & {s_k }  \\
   1 &  \cdots  & k  \\
 \end{array} } \right)} &
 , &
 {\gamma _i^{ - 1}  = \left( {\begin{array}{*{20}c}
   1 & {s_1 }  \\
    \vdots  &  \vdots   \\
   k & {s_k }  \\
 \end{array} } \right)}
\end{array}
\]
where
\[
\begin{array}{*{20}c}
   {\{ s_1 , \cdots ,s_k \}  \subset \{ 1, \cdots ,n\} } & {} & {i \ne j \to s_i  \ne s_j }
\end{array}.
\]
\[
\begin{array}{*{20}c}
   {B = \gamma _i A,} & {B = A\gamma _i^T ,} & {B = \gamma _i A\gamma _i^T ,} & {B = \gamma _i^{ - 1} A}.
\end{array}
\]

(1) If $A$ is a column vector or matrix,
$B=\gamma_i A$
assigns rows $s_1,\dots,s_k$ of $B$
to rows $1,\dots,k$ of $A$.

(2) If $A$ is a row vector or matrix,
$B=A\gamma_i^T$
assigns columns $s_1,\dots,s_k$ of $B$
to columns $1,\dots,k$ of $A$.

(3) If $A$ is a matrix,
$B=\gamma_i A\gamma_i^T$
fills the submatrix of $B$
indexed by $s_1,\dots,s_k$
with the corresponding values of $A$.

(4) If $A$ is a column vector or matrix,
$B=\gamma_i^{-1}A$
projects rows $s_1,\dots,s_k$ of $A$
to rows $1,\dots,k$ of $B$.

\subsection{Principle of Reorganization}

\subsubsection{Structural Principle}
\begin{figure}[H]
  \centering
  \includegraphics[width=0.55\textwidth]{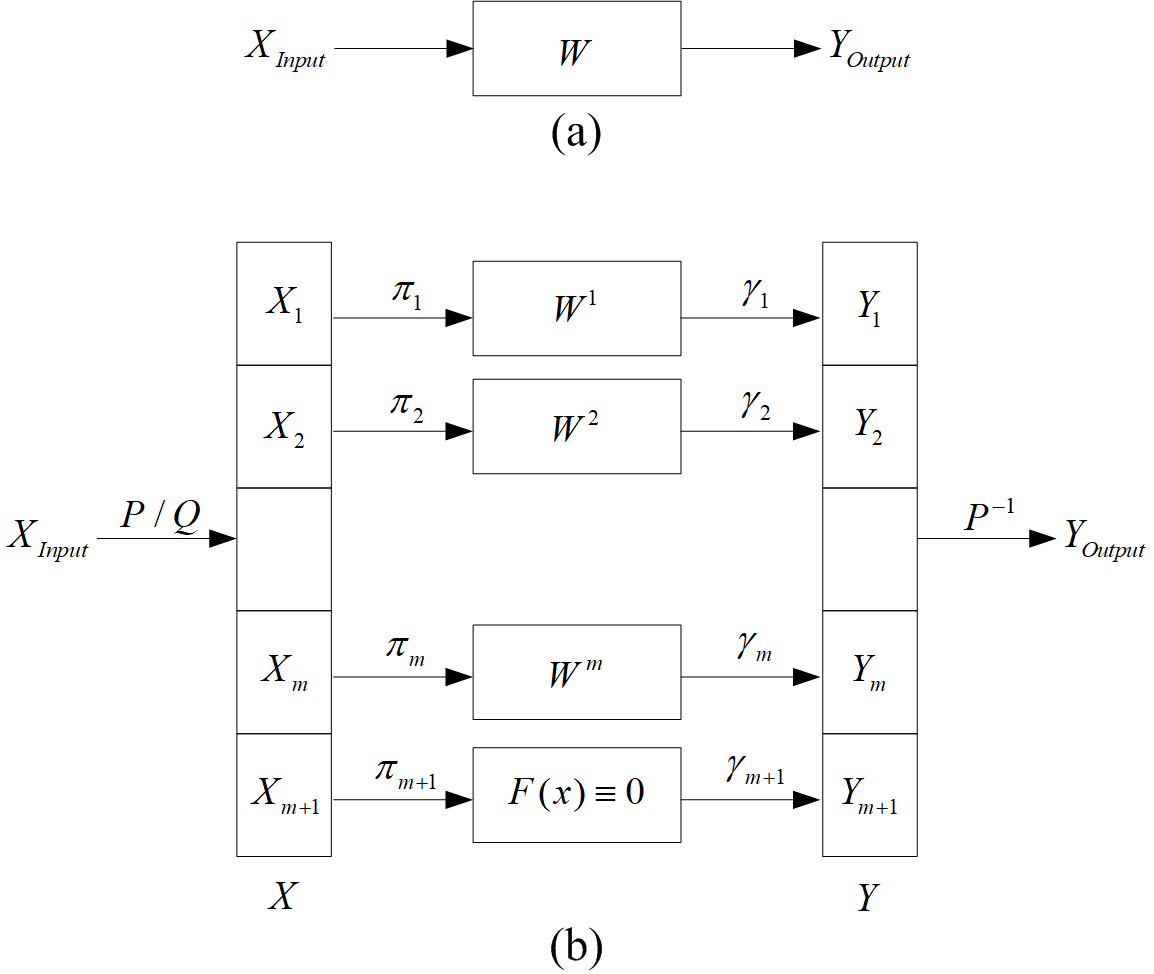}
  \caption{Structural principle of inference-system reorganization}
  \label{fig:inference-restructuring}
\end{figure}

Figure~\ref{fig:inference-restructuring} illustrates the basic principle
of inference-system structural reorganization.
Panel (a) shows the original inference system before reorganization,
while panel (b) shows an equivalent reorganized representation
obtained without changing the functionality of the system.
The entire reorganization process consists of three structural stages.

\paragraph{ (I) Input decomposition stage. }

The original input vector $X_{\mathrm{Input}}$
is first mapped to an intermediate representation $X$
by reordering its component positions
through the permutation $P$ in (\ref{PWPT})
or the permutation $Q$ in (\ref{EQ:FNN-1});
subsequently,
a family of projections
\[
\pi_1,\dots,\pi_m,\pi_{m+1}
\]
projects the subvectors $X_i$ of $X$
onto the input sides of $W^i$.

We now clarify what $X$ is.
Let the operator part of the original inference system be
\begin{equation}\label{EQ:a}
Y_{Output}  = W  X_{Input}.
\end{equation}
If $P$ or $Q$ is a permutation, then
\begin{equation}\label{EQ:b}
\begin{array}{*{20}c}
   {\left( {PY_{Output} } \right) = \left( {PWP^T } \right)\left( {PX_{Input} } \right)} & {or} & {\left( {PY_{Output} } \right) = \left( {PWQ^T } \right)\left( {Q X_{Input} } \right)}
 \end{array}
\end{equation}
is a computational system equivalent to (\ref{EQ:a}).
Accordingly,
\[
\begin{array}{*{20}c}
   {X = PX_{Input} } & {or} & {X = QX_{Input} }
 \end{array}
\]
is the input corresponding to the equivalent operator
\[
\begin{array}{*{20}c}
   {PWP^T } & {or} & {PWQ^T }.
 \end{array}
\]

The subvectors
\[
X = (X_1 , \cdots ,X_m ,X_{m + 1} )^T
\]
namely
$$
X_i(i=1,\ldots,m+1),
$$
directly correspond to the input parts
of the independent sub-operators
$$
W^i.
$$
To convert these subvectors of $X$
into the actual input vectors
of the parallel sub-operators $W^i$,
the projection $\pi_i$ serves precisely this role.

This stage involves only
the positional reordering of input-vector components
and subset mappings.
It contains no numerical computation,
is fully compatible with the original input interface of the system,
and does not alter the semantics of the state variables.

\paragraph{  (II) Parallel sub-operator stage. }

Each sub-input $X_i$
acts on its corresponding parallel sub-operator $W^{i}$,
producing an intermediate output
\[
Y_i = W^{i} X_i,
\qquad i=1,\dots,m.
\]
These sub-operators are structurally independent,
correspond to the independent sub-operators
in the parameter matrix,
and can be executed in parallel at the computational level.

\paragraph{  (III) Output embedding and inverse-permutation stage.}

The outputs of the channels
$Y_1,\dots,Y_m,Y_{m+1}$
are embedded into $Y$
through a family of pairwise disjoint embedding mappings
\begin{equation}\label{EQ:gamma}
\begin{array}{*{20}c}
   {\gamma _1 , \ldots ,\gamma _m ,\gamma _{m + 1} } & {} &
   {i \ne j \;\Rightarrow\;
   \mathrm{Im}(\gamma_i)\cap\mathrm{Im}(\gamma_j)=\emptyset}
\end{array}
\end{equation}
and are then mapped back
to the original component coordinate system
through the inverse permutation $P^{-1}$,
yielding the final output $Y_{Output}$.

Similar to the positional reordering and decomposition
of the input variables,
the outputs $Y_i$ of the independent sub-operators
are first embedded through $\gamma_i$
into the corresponding positions
of the output variable
\[
\begin{array}{*{20}c}
   {PW  P^T } & {or} & {PW  Q^T }
 \end{array}
\]
namely $Y=PY_{Output}$,
and are then transformed into
the output of the original system
through
\[
Y_{Output}  = P^{ - 1} Y.
\]

This stage involves only
the aggregation of output-vector components
and the positional reordering of the output vector.
It contains no numerical computation,
is fully compatible with the original output interface of the system,
and does not alter the semantics of the state variables.

\paragraph{  Dormant neurons.}

After annealing and structural reorganization are completed,
the graph corresponding to the parameter matrix
may contain weakly connected subgraphs
consisting of only a single node
(i.e., isolated points).
In matrix representation,
such isolated points correspond to rows or columns equal to zero
(or both row and column equal to zero
in the case of artificial neural networks with feedback connections).

From the operator perspective,
such neurons either have no effective response to any input,
so that their output is identically the zero map;
or their input is identically zero
(or both their input and output are identically zero
in the case of artificial neural networks with feedback connections).
We refer to such structures as \emph{dormant neurons}.

In the reorganized system,
all dormant neurons are uniformly collected
into a single zero-operator channel:
\[
Y_{m+1} = F(X_{m+1}) \equiv 0.
\]
This work does not recommend
immediately deleting this channel at this stage;
instead,
it is explicitly retained as a placeholder module.
On the one hand,
this preserves the consistency and reversibility
of the state variables;
on the other hand,
the redundancy introduced by this choice
is only of order $O(n)$,
which is negligible in engineering implementation.

\subsubsection{Characteristics of the Structural Principle}

\paragraph{ (1) Functional equivalence.}

Structural reorganization does not introduce any new inference capability
and does not alter the functionality of the overall operator.
Under appropriate permutation transformations
$P$ or $(P,Q)$,
the systems before and after reorganization
are strictly equivalent in the operator sense,
that is,
they produce identical outputs for any given input.

\paragraph{ (2) Invariance of input–output format.}

Both the original system and the reorganized system
use the same input vector
$X_{\mathrm{Input}}$
and output vector
$Y_{\mathrm{Output}}$.
Their dimensionality,
component order,
and semantic interpretation remain unchanged.
Consequently,
the reorganized system can replace the original one
directly without modifying any external calling interfaces.

\paragraph{ (3) Parallelism of permutation / projection / embedding. }

In this work,
the permutation operator $P$,
projection operators $\pi_i$,
and embedding operators $\gamma_i$
are implemented as bijective functions
defined on finite index sets:
\[
\begin{array}{*{20}c}
   {P:\{ 1, \ldots ,n\}  \to \{ 1, \ldots ,n\} }  \\
   {\pi _i ,\gamma _i :\{ 1, \cdots ,k\}  \to \{ s_1 , \cdots ,s_k \}  \subset \{ 1, \ldots ,n\} }
 \end{array}
\]

Under this implementation,
permutations ($P,Q$),
projections ($\pi_i$),
and embeddings ($\gamma_i$)
are essentially operations of index reordering,
component extraction,
and component embedding.
They involve no multiplication or addition operations,
and the access cost for reordering,
projection,
and embedding
is $O(n)$.

Moreover,
$\pi_i(i=1,\ldots,m+1)$
and
$\gamma_i(i=1,\ldots,m+1)$
have no storage conflicts with each other,
and therefore can theoretically be executed in parallel.
Similarly,
the transformations
$X=PX_{Input}$
and
$Y_{Output}=P^{-1}Y$
can also be executed in parallel
with appropriate segmentation granularity.

\paragraph{ (4) Runtime cost.}

If the reorganized system can be divided
into $k$ independent execution modules,
then the computational cost of the corresponding part
of the reorganized system
can be reduced to approximately $1/k$
of the original cost,
while the required scale of computing hardware
is reduced proportionally.

Inference restructuring decomposes a single monolithic computation
into independent sub-computations that can be executed in parallel.
The entire process introduces no new computational semantics,
does not alter the model interface,
and achieves equivalent inference
solely through input–output index mappings.

\subsection{Reorganization of Multilayer Systems}
\begin{figure}[H]
  \centering
  \includegraphics[width=1\textwidth]{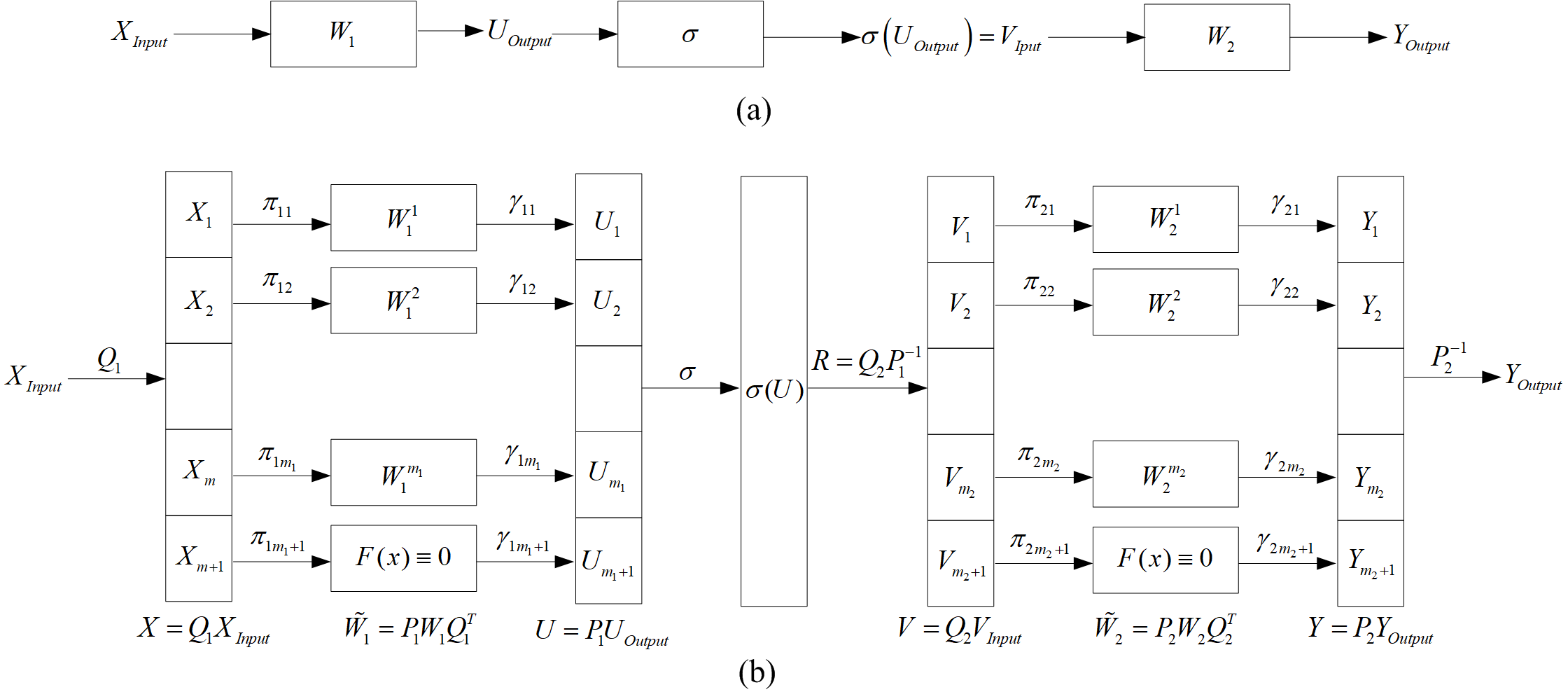}
  \caption{Structural principle of multilayer system reorganization}
  \label{fig:multilayer-reorganization-detailed}
\end{figure}

Figure~\ref{fig:multilayer-reorganization-detailed}
illustrates the structural reorganization process
of a two-layer inference system
while preserving the overall computational semantics
and the input–output interface.

(a) shows the original computational flow.

The input $X_{\text{Input}}$
is first processed by the first-layer linear operator $W_1$
to generate the intermediate output $U_{\text{Output}}$.
A component-wise nonlinear mapping $\sigma(\cdot)$
is then applied,
yielding
$\sigma(U_{\text{Output}})=V_{\text{Input}}$.
This vector serves as the input
to the second-layer linear operator $W_2$,
which finally produces the output
$Y_{\text{Output}}$.

(b) shows the corresponding structured reorganization form.

First,
the input permutation mapping $Q_1$
reorders the input vector
\[
X = Q_1 X_{\text{Input}},
\]
and the first-layer operator is represented as
\[
\tilde{W}_1 = P_1 W_1 Q_1^{T},
\]
so that under the reordered coordinate system
it decomposes into a set of sub-operators $W_1^i$.

Each sub-operator acts on the corresponding input component $X_i$
and produces an intermediate state $U_i$.
These intermediate outputs are collectively written as
\[
U = P_1 U_{\text{Output}}.
\]

Next,
the nonlinear mapping $\sigma(\cdot)$
is applied component-wise to $U$.
Through the reordering mapping
\[
R = Q_2 P_1^{-1},
\]
the vector $\sigma(U)$
is reorganized as the second-layer input
\[
V = Q_2 V_{\text{Input}}.
\]

Under this representation,
the second-layer operator is written as
\[
\tilde{W}_2 = P_2 W_2 Q_2^{T},
\]
and decomposed into sub-operators $W_2^j$,
each acting on $V_j$
to generate the output component $Y_j$.

Some channels correspond to the zero operator
$F(x)\equiv 0$,
indicating subpaths that do not participate
in effective computation under this structural representation.

Finally,
the output is recombined through the inverse permutation
$P_2^{-1}$
to obtain
$Y_{\text{Output}}$,
which remains consistent with the original system
in terms of overall computational semantics
and interface behavior.

In artificial neural networks without feedback structure,
the parameter matrix is viewed as a linear operator
mapping one class of neuron states to another.
The corresponding graph structure is therefore
a bipartite graph rather than a simple directed graph.

Consider a two-layer feed-forward inference system as an example.
The first-layer operator maps the input space $\mathbb{R}^{d}$
to an intermediate space $\mathbb{R}^{k}$,
and the second-layer operator maps $\mathbb{R}^{k}$
back to the output space $\mathbb{R}^{d}$,
where $k$ may take a value such as $4d$
or another dimension different from $d$.

Therefore,
the structural principle illustrated in
Figure~\ref{fig:multilayer-reorganization-detailed}
can in principle be directly applied
to the reorganization of a typical
$d\!-\!4d\!-\!d$ feed-forward network.
However,
this is not an optimal solution,
because the $d\!-\!4d\!-\!d$ feed-forward network itself
already possesses a block-based multi-channel parallel structure.
\[
\begin{array}{*{20}c}
   {U_{Output}  = W_1 X_{Input} } & {\Leftrightarrow } &
   {\left( {\begin{array}{*{20}c}
   {U_{Output}^1 }  \\
    \vdots   \\
   {U_{Output}^4 }  \\
 \end{array} } \right)}
\end{array}
=
\left( {\begin{array}{*{20}c}
   {W_1^1 } & {} & {}  \\
   {} &  \ddots  & {}  \\
   {} & {} & {W_1^4 }  \\
 \end{array} } \right)
\left( {\begin{array}{*{20}c}
   {X_{Input} }  \\
    \vdots   \\
   {X_{Input} }  \\
 \end{array} } \right)
\]
\[
\begin{array}{*{20}c}
   {Y_{Output}  = W_2 V_{Input} } &  \Leftrightarrow  &
   {Y_{Output}  =
   \left( {\begin{array}{*{20}c}
   {W_2^1 } &  \cdots  & {W_2^4 }
 \end{array} } \right)
 \left( {\begin{array}{*{20}c}
   {V_{Input}^1 }  \\
    \vdots   \\
   {V_{Input}^4 }
 \end{array} } \right)
 =
 \sum\limits_{q = 1}^4 {W_2^q } V_{Input}^q }
\end{array}
\]

\begin{figure}[H]
  \centering
  \includegraphics[width=0.95\textwidth]{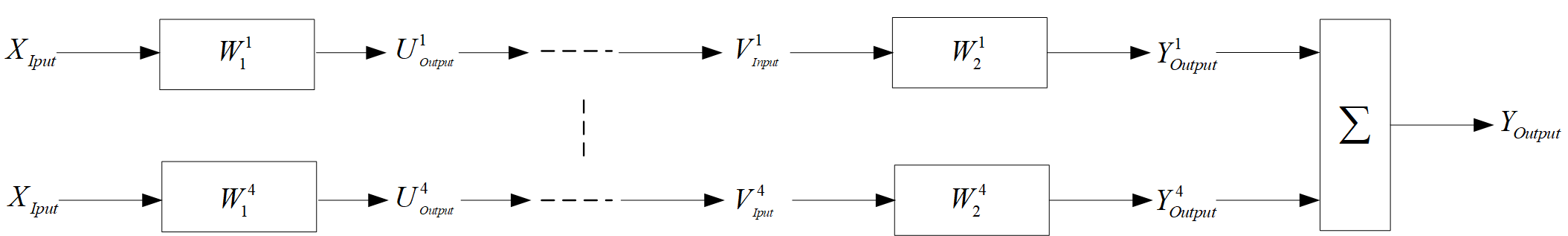}
  \caption{Multi-channel structure of a $d\!-\!4d\!-\!d$ feed-forward network}
  \label{fig:FNN:d-4d-d}
\end{figure}

In order not to destroy its original
multi-channel parallel property,
the $d\!-\!4d\!-\!d$ feed-forward network
should be reorganized layer by layer
along its channels,
as illustrated in
Figure~\ref{fig:FNN:d-4d-d}.

\subsection{A Three-Stage Framework for Inference-System Restructuring}

From the perspectives of engineering implementation
and system evolution,
the structural restructuring of an inference system
is best organized as a staged process:
(1) a validation stage;
(2) an operational stage;
and (3) an upgrade stage
(improvement, expansion, and continued learning).

\paragraph{ 1. Validation stage (trial-run stage).}

The central task of the validation stage
is to verify the effectiveness and consistency
between the restructured system
and the original system.

Its necessity arises from the fact that
the structure obtained after annealing
depends on the statistical significance level.
Different confidence levels
may induce different edge-removal sets
and therefore different structural decompositions.
Consequently,
comparative testing between the new and the original systems
must be carried out under representative confidence levels.

The validation procedure follows a unified principle:
the original system and the restructured system
are executed in parallel under identical inputs,
and their outputs are compared.

Validation data should preferably be drawn
from the validation set used in the late stage of training;
when necessary,
new data not involved in training
may also be introduced
to assess generalization effects.

The focus of comparison is not strict equality
at the level of individual samples or components,
but rather whether the outputs of key tasks remain consistent,
whether the deviation stays within acceptable bounds,
and whether the results vary continuously
and remain interpretable
as the confidence level changes.

From the engineering perspective,
the validation stage may directly adopt
the structural scheme illustrated in
Figure~\ref{fig:inference-restructuring}:
\[
\text{permutation}
\rightarrow
\text{projection}
\rightarrow
\text{parallel sub-operators}
\rightarrow
\text{embedding}
\rightarrow
\text{inverse permutation}.
\]

This approach maximally isolates
the effects of ``structural change''
from those of ``scheduling optimization,''
while preserving system reversibility.
At this stage,
optimal parallel efficiency is not the primary concern;
reduced matching efficiency caused by
asynchronous module processing rates is acceptable,
since the purpose of this stage
is to verify the correctness and controllability
of the structural restructuring.

\paragraph{ 2. Engineering implementation in the operational stage:
a transform–inference two-tier service architecture. }

After validation is completed
and the structure is fixed,
the system enters the operational stage.

The primary efficiency issue at this stage
arises from differences in computational scale
among parallel submodules.
Such differences lead to inconsistent processing rates
across modules,
thereby limiting overall parallel efficiency.

Without destroying the existing advantages
of structural restructuring
(functional equivalence,
unchanged interfaces,
and low index-operation overhead),
we recommend adopting a
\emph{transform–inference two-tier service architecture}.

In this architecture,
the transform layer is responsible for
all structural operations,
including

permutations $P,Q,P^{-1}$,
projections $\{\pi_i\}$,
submodule triggering and scheduling,
and embeddings $\{\gamma_i\}$.

The inference layer deploys each sub-operator $W_\sigma^{i}$
as an independent service
that performs only the numerical inference
\[
Y_i = W_\sigma^{i}X_i .
\]

Under this architecture,
the transform layer centrally manages
the structural and scheduling logic,
while inference-layer modules
can elastically scale
according to their computational loads.

Modules with larger computational requirements
can be allocated more instances,
while those with smaller workloads
can operate with fewer instances.
This significantly reduces idle resource waste
and improves overall resource utilization.

Importantly,
this implementation does not alter
the mathematical structure of any operator;
it improves engineering efficiency
purely through runtime scheduling.

\paragraph{ 3. Upgrade learning after restructuring:
a two-step matrix-update perspective.}

The restructured inference system
can still support upgrade learning
without expanding the state variables.

Consider a single-layer linear operator.
Let the forward propagation be
\[
Y = AX,
\]

which corresponds to the linear mapping
$Y(t)=A(t)X(t)$.
Under gradient descent,
the standard backpropagation update
can be derived directly
from the chain rule
\cite{goodfellow2016deep,rumelhart1986learning}:
\[
A(t+1)
=
A(t)
-
\eta\, G_{Y(t)} X^T(t),
\qquad
G_{Y(t)} = \frac{\partial L}{\partial Y(t)}.
\]

This update has a clear two-step structure:
\[
\Delta A(t) = \eta\, G_{Y(t)} X^T(t),
\qquad
A(t+1) = A(t) - \Delta A(t).
\]

Therefore,
as long as the input $X$
and the output $Y$
remain identical before and after restructuring,
the learning-update mechanism
remains algebraically unchanged.
Structural restructuring modifies
only the organization of operators,
not the update rule itself.

In the restructured system,
the global update
$\Delta W(t)$
can be redistributed to each sub-operator
through permutations,
inverse embeddings,
and projections:
\begin{equation}\label{return}
\begin{array}{*{20}c}
   {W_i (t + 1) = W_i (t) - \gamma _i^{ - 1}
   \left( {P\,\Delta W(t)\,P^T } \right)\pi _i^T }  \\
   {or}  \\
   {W_i (t + 1) = W_i (t) - \gamma _i^{ - 1}
   \left( {P\,\Delta W(t)\,Q^T } \right)\pi _i^T }
\end{array}
\end{equation}
where
\[
\Delta W(t)
=
\eta\,
G_{Y_{\mathrm{Output}}(t)}
X_{\mathrm{Input}}^T(t),
\qquad
G_{Y_{\mathrm{Output}}(t)}
=
\frac{\partial L}{\partial Y_{\mathrm{Output}}(t)}.
\]

Here,
$\gamma_i^{-1}$ denotes the inverse (projection)
of the embedding $\gamma_i$.
Under the joint action of $\gamma_i^{-1}$
and $\pi_i^T$,
the corresponding submatrix
of the global update matrix
\[
\begin{array}{*{20}c}
   {P\,\Delta W(t)\,P^T } & {or} & {P\,\Delta W(t)\,Q^T }
\end{array}
\]
is extracted
and assigned to the corresponding sub-operator.

\section{Conclusion}

From the perspective of system and engineering design,
this work shows that large-scale artificial neural systems,
after training is completed,
are not indivisible black-box entities.
At the parameter level,
the dynamics of learning have already
naturally deposited a number of stable substructures.
These structures are not introduced by external design rules;
rather,
they are statistical outcomes
formed through long-term iterations
of gradient reinforcement and suppression.

Accordingly,
the structural restructuring proposed in this paper
should not be interpreted as a new model-construction method,
but rather as a form of \emph{post-training static analysis}.
At the deployment stage of a model,
there is no need to revisit the training trajectory
or reconstruct the history of gradient updates.
Using only the parameter matrix obtained after training,
one can statistically distinguish
between dependencies that have been confirmed
through the learning process
and parameter components that have not developed
structural influence.
The structure is not designed;
it is revealed.

The graph structure that emerges after annealing
naturally induces,
in the graph-theoretic sense,
a modular partition
together with an ordering of dependencies.
An inference system that originally operates
as a monolithic operator
can therefore be reorganized,
without changing its input--output interface
or its overall computational semantics,
into several relatively self-contained
sub-operator systems.
Such a decomposition provides
a direct engineering pathway
for block-wise execution,
module-level parallel scheduling,
and progressive inference
under resource-constrained environments.

As model sizes continue to grow,
the system bottleneck
is gradually shifting
from the sheer number of parameters
to the form of structural coupling
and the efficiency of scheduling.
If the structural differentiation
already formed during training
is ignored,
simply enlarging the model scale
will inevitably accumulate system complexity.
By explicitly identifying and consolidating
the effective structures that emerge during training
in the post-training stage,
one can improve system scalability,
maintainability,
and long-term evolutionary potential
without sacrificing model capability.

Therefore,
this work does not propose
a new model architecture.
Instead,
it presents a structural treatment principle
for existing models:
elevating the implicitly formed structures
of the learning process
into system-level objects,
so that the design focus of inference systems
shifts from sheer scale expansion
toward structural organization
and sustainable operation.

\part{Matrix-Based Structural Decomposition Algorithm}

This part addresses the operational problem of the present work:
the structural decomposition of an artificial neural network
or its parameter matrix.

For the annealed parameter matrix $W$,
we need to determine row and column permutations
(reindexing)
$P,Q$
so that the rows and columns of $W$
(or of its Boolean adjacency matrix $M$)
can be reordered,
thereby revealing a block-diagonal structure.

The computation of matrix row--column permutations
may be carried out using graph-search methods.
In this section,
we present a solution algorithm based purely on matrix operations.

While,
compared with classical graph-search algorithms,
its total workload under the traditional RAM model
is not necessarily superior,
in the context of GPU-based matrix parallel computation
its iteration dependency chain
(the critical path or serial depth)
does not exceed
$O(\lceil \log_2 n\rceil)$,
which makes the method practically feasible.

Traditional graph-search algorithms
(such as SCC decomposition based on DFS)
have a total workload of $O(n+m)$
under the RAM model,
where $n=|V|$ denotes the number of vertices
and $m=|E|$ denotes the number of edges.
However,
using graph-search algorithms
typically requires converting $W$
into an explicit graph representation
(such as an edge list or sparse format).
This conversion itself requires
$O(n^2)$ work
and involves a large amount
of irregular memory access.

Considering the hardware configurations
and parallel-computation characteristics
commonly found in AI service infrastructures,
matrix-based algorithms
remain a practical option.

The main advantage of the matrix algorithm
is that it does not require changing
the representation paradigm of the parameter matrix.
Instead,
one directly constructs
a Boolean relation matrix
or adjacency matrix $M$
from $W$,
and obtains the permutations $P,Q$
through matrix operations on $M$.
These permutations can then be applied
to reorder the rows and columns
of $M$ (or $W$),
allowing the structural form
of the inference system
to be observed under different annealing
confidence levels.

This property makes the method convenient
for repeated computation
under different annealing confidence levels,
thereby helping identify
a relatively reasonable annealing threshold.
If applied during the learning stage of a model,
it can further serve
as a monitor for the formation
and evolution of system structure.

This part may also serve
as a theoretical reference
for the design of more efficient algorithms.

\section{Mathematical Background}\label{subsec:math-background}

This section summarizes the terminology and mathematical background
required by the algorithm.
These concepts constitute the basic theoretical foundation
for the construction of the algorithm.

Definition~\ref{defn1} and Theorem~\ref{thm1}
collect the background knowledge used in this work,
including the Boolean matrix representation of graphs,
Boolean matrix operations,
and transitive closure operations.
These topics belong to the standard material
of the relevant disciplines
and can be found in textbooks or monographs
in the corresponding fields.
\[
G = \langle V,\, E \subseteq V\times V,\, A\rangle
\]
denotes a simple directed graph,
where $n=|V|$ is called the order of $G$.
Between any pair of vertices in a simple directed graph
there exists at most one directed edge,
making it suitable for representing relations.
If $(v_i,v_j)\in E$ denotes a directed edge in $G$,
then the relation between $G$ and $A$ is
\[
\left\{ {\begin{array}{*{20}c}
   {(v_i ,v_j ) \in E} & {} & {A(i,j) = 1,}  \\
   {(v_i ,v_j ) \notin E} & {} & {A(i,j) = 0.}
 \end{array} } \right.
\]
The matrix $A$ is called the adjacency matrix of $G$.

\medskip
\noindent
\textbf{Notation.}
For any matrix $A$,
$A(i,j)$ denotes the element in the $i$-th row and $j$-th column,
$A(i,:)$ denotes the $i$-th row vector,
and $A(:,j)$ denotes the $j$-th column vector.

\begin{definition}\label{defn1}\cite{kleene1956representation,warshall1962boolean,kim1982boolean}
Boolean product (power) of Boolean matrices,
transitive closure of Boolean (adjacency) matrices,
and the matrix representation of the underlying graph.
\begin{enumerate}

\item
Let $A \in\{0,1\}^{m \times n}$ and
$B \in\{0,1\}^{n \times l}$ be Boolean matrices.
If $C=A\odot B$,
then $C$ is called the Boolean product of $A$ and $B$,
whose elements are defined by
\[
C(i,j)=\bigvee_{k=1}^{n}\bigl(A(i,k)\wedge B(k,j)\bigr),
\qquad i=1, \dots m ; j=1,\dots,l.
\]
Furthermore,
for $A \in\{0,1\}^{n \times n}$,
define $A^k=A^{k-1}\odot A$
as the $k$-th power of $A$.

\item
\cite{kleene1956representation}
For a Boolean matrix $A\in\{0,1\}^{n\times n}$,
$A^+$ and $A^*$ are called the transitive $+$ closure
and the transitive $*$ closure of $A$
(respectively),
where $I$ denotes the Boolean identity matrix:
\[
A^+=\bigvee_{k=1}^{n-1}A^k
= A\vee A^2\vee\cdots\vee A^{n-1},
\]
\[
A^*=I\vee A^+
=\bigvee_{k=0}^{n-1}A^k
= I\vee A\vee A^2\vee\cdots\vee A^{n-1}.
\]

\item
If $A$ is the adjacency matrix
of a simple directed graph $G=<V,E>$,
and $D=A\vee A^T$,
then $D$ is the adjacency matrix
of the underlying undirected graph of $G$, namely
\[
D(i,j)=1 \iff (v_i,v_j)\in E\ \vee\ (v_j,v_i)\in E .
\]

\end{enumerate}
\end{definition}

\begin{theorem}\label{thm1}

Let $A$ be the adjacency matrix
of a simple directed graph $G=\langle V,E\rangle$,
and let $A^+$ and $A^*$ denote
the transitive $+$ closure and transitive $*$ closure of $A$.

\begin{enumerate}

\item
Matrix representation of path existence between vertices:
\[
A^+(i,j)=1 \iff A^*(i,j)=1
\iff
\text{there exists a path from } v_j \text{ to } v_i .
\]
Compared with $A^+$,
$A^*$ additionally includes reflexivity,
meaning that a path from $v_i$ to itself is assumed.

\item
(Kleene--Valiant) \cite{valiant1975general}

Let $A^{(0)}=I\vee A$,
and define recursively
\[
A^{(t+1)}=A^{(t)}\odot A^{(t)},
\qquad t=0,1,2,\dots
\]
Then when
$t=\left\lceil \log_2(n-1)\right\rceil$
or when $A^{(t+1)}=A^{(t)}$,
we obtain $A^{(t)}=A^*$.

\item
Let
\[
B = A^* \wedge (A^*)^T .
\]
Then $B$ is the matrix representation
of the mutual reachability relation
between vertices in $G$:
$B(i,j)=1$ if and only if
$v_i$ and $v_j$ are mutually reachable.

This mutual-reachability relation
induces a partition on $V$.
Each equivalence class corresponds
to a maximal strongly connected subgraph
(strongly connected component, SCC),
and each vertex belongs to exactly one SCC.

\item
Let
\[
D=A\vee A^T
\]
be the adjacency matrix of the underlying graph of $G$.
Then $D^*$ also induces a partition of $V$:
each equivalence class corresponds
to a maximal weakly connected subgraph,
and no path exists in either direction
between different equivalence classes.

\end{enumerate}

\end{theorem}

\begin{proof} 

(1)
A basic theorem in graph theory
and the theory of transitive closure.
It corresponds to the graph-theoretic result
that if a path exists between two vertices,
then there exists a path of length
at most $n-1$.

Proof idea:
if there exists a path from $v_p$ to $v_q$
whose length exceeds $n-1$,
then the path must contain repeated vertices.
Removing the intermediate segment between repeated vertices
yields a path whose length does not exceed $n-1$.

(2)
Boolean matrix polynomials satisfy the following property:
\[
\left(\bigvee_{p=0}^{k}A^p\right)^2
=
\bigvee_{p=0}^{2k}A^p.
\]

That is,
the Boolean square ($\odot$)
of a complete unary polynomial of degree $k$
containing the identity element $A^0=I$
produces a complete unary polynomial
of degree $2k$
that also contains the identity element.

Since $A^*$ is a complete unary polynomial
of degree $n-1$,
its iterative computation
can be completed within
$t=\left\lceil \log_2(n-1)\right\rceil$ iterations.

Moreover,
when $A^{(t+1)}=A^{(t)}$,
subsequent iterations no longer change $A^{(t)}$,
which may also serve as a termination condition.

(3)
The relation corresponding to
$B=A^*\wedge (A^*)^T$
inherits the reflexivity and transitivity of $A^*$,
while the conjunction with its transpose
adds symmetry.
Therefore $B$ represents an equivalence relation.

An equivalence relation induces a partition.
Vertices in the same equivalence class
are mutually reachable
and cannot be further enlarged;
thus each equivalence class corresponds
to a maximal strongly connected subgraph
(SCC).

(4)
The symmetry of $D=A\vee A^T$
is inherited by $D^*$,
while $D^*$ also possesses reflexivity
and transitivity.
Hence $D^*$ is likewise
the matrix representation
of an equivalence relation,
which also induces a partition:
the partition of the underlying graph of the original graph;
consequently,
it is also the partition
into weakly connected subgraphs
of the original graph.

\end{proof}

\section{Structural Decomposition of Bipartite Networks}

This section corresponds to the structural decomposition
of parameter matrices in feedforward neural networks.

\subsection{Method Principle}

The parameter matrix of a feedforward neural network
does not encode state feedback loops.
Its structure corresponds to a bipartite graph
$X\to Y$ rather than a general directed graph.
To characterize this structure,
we analyze the Boolean equivalent form of
\[
Y = WX
\]
given by
\[
Y = MX .
\]

The matrix $M$ is regarded as the matrix representation
of the mapping relation $E$
in the bipartite graph
\[
G=\langle Y\cup X,\,E\subset Y\times X\rangle .
\]

The goal is to determine the partition of $G$
into maximal weakly connected subgraphs.

The basic idea is as follows.
If $Y_s\cup X_s$ forms a maximal weakly connected subgraph of $G$,
then the components in $Y_s$
are connected through their shared sources in $X_s$.
That is,
if two image nodes $y_i,y_j$
share a common source node $x_k$,
then they belong to the same connected subgraph.

Accordingly,
a co-source relation ${\cal Y}$
can be constructed on $Y$.
Taking the transitive $*$ closure of this relation
produces an equivalence-class partition of $Y$.
For each equivalence class $Y_s$,
its source set
\[
\operatorname{Dom}(Y_s)
\]
is then determined.
Thus
\[
\operatorname{Dom}(Y_s) \to Y_s
\]
forms a maximal weakly connected subgraph of $G$.

Theorem~\ref{THM:co-Dom}
provides a method for generating the equivalence-class partition.
Theorem~\ref{THM:BG-compression}
compresses the partition into two indicator matrices,
from which the permutations $P,Q$
can be directly constructed.
The permutations $P,Q$ transform $M$
into a block-diagonal structure.

Similarly,
one may construct a co-image relation on $X$
to complete the partition.
When $M$ is square,
the two constructions are essentially equivalent.
When $M$ is rectangular,
the two directions may differ
in computational scale and complexity.

In this work,
we present the algorithm based on the co-source relation;
the construction based on the co-image relation is analogous.

\begin{definition}\label{DF:BG} Bipartite Graph and Co-source Relation

\begin{enumerate}

\item
A graph $G=\langle Y\cup X,\,E\subset Y\times X\rangle$
is called a bipartite graph,
where $X$ is the set of source nodes
and $Y$ is the set of image nodes.
$(y_i,x_j)\in E$ denotes a directed edge
from $x_j$ to $y_i$.

\item
Given a parameter matrix
$W\in\mathbb{R}^{m\times n}$,
let $M\in\{0,1\}^{m\times n}$ be the
$P$-mapping relation matrix of $W$,
defined by
\[
M(i,j)=
\begin{cases}
1, & P\!\bigl(W(i,j)\bigr),\\
0, & \neg P\!\bigl(W(i,j)\bigr),
\end{cases}
\qquad i=1,\dots,m,\ j=1,\dots,n.
\]
Here $P(x)$ indicates that $x$
satisfies property $P$.
The matrix $M$ is the matrix representation
of the mapping relation $E$,
and thus the bipartite graph may also be written as
\[
G=\langle Y\cup X,\,E\subset Y\times X,\,M\rangle .
\]

\item \cite{ZweigKaufmann2011}
The co-source relation on the image node set $Y$
is defined as
${\cal Y}\subset Y\times Y$:
\[
(y_i,y_j)\in{\cal Y}
\ \Leftrightarrow\ 
\exists\,x_k\in X\ 
\bigl[(y_i,x_k)\in E \wedge (y_j,x_k)\in E\bigr].
\]

\end{enumerate}

\end{definition}

\noindent
\textbf{Remark.}
The matrix $M$ in Definition~\ref{DF:BG}
is not the adjacency matrix of the bipartite graph.
Instead,
it represents the mapping relation
$E\subset Y\times X$.

In fact,
the adjacency matrix of the bipartite graph $G$
can be written as
\[
\left(
\begin{array}{cc}
0 & 0\\
M & 0
\end{array}
\right).
\]

\begin{theorem}\label{THM:co-Dom}

Let
$G=\langle Y\cup X,\,E\subset Y\times X,\,M\rangle$
be a bipartite graph,
where $|Y|=m$, $|X|=n$,
and $M\in\{0,1\}^{m\times n}$.
Then

\begin{enumerate}

\item
The matrix representation of the co-source relation
${\cal Y}$ is
\[
M_Y=M\odot M^T\in\{0,1\}^{m\times m},
\]
namely, for any $i,j$,
\[
M_Y(i,j)=1
\ \Leftrightarrow\ 
(y_i,y_j)\in{\cal Y}
\ \Leftrightarrow\ 
\exists\,x_k\in X
\bigl[(y_i,x_k)\in E\wedge (y_j,x_k)\in E\bigr].
\]

\item
Let $B_Y=M_Y^*$ be the transitive $*$ closure of $M_Y$.
Then $B_Y$ is the matrix representation
of an equivalence relation on $Y$,
which induces a partition
\begin{equation}\label{EQ:YD}
\bar Y=\{Y_1,\dots,Y_{k}\},\quad
Y_p\cap Y_q=\emptyset\ (p\ne q),\quad
\bigcup_{p=1}^{k}Y_p=Y.
\end{equation}
Moreover,
the source sets of different equivalence classes
are mutually disjoint:
\[
\operatorname{Dom}(Y_p)\cap \operatorname{Dom}(Y_q)=\emptyset
\quad (p\ne q).
\]

\end{enumerate}

\end{theorem}

\begin{proof}

Let
$Y=\{y_1,\dots,y_m\}$,
$X=\{x_1,\dots,x_n\}$,
and $M\in\{0,1\}^{m\times n}$
be the mapping relation matrix:
\[
M(i,j)=1 \iff (y_i,x_j)\in E,
\qquad i=1,\dots,m,\ j=1,\dots,n.
\]

\medskip
\noindent
\textbf{(1)}

For any $i,j\in\{1,\dots,m\}$,
\[
M_Y(i,j)=(M\odot M^T)(i,j)
=
\bigvee_{k=1}^{n}\bigl(M(i,k)\wedge M^T(k,j)\bigr)
=
\bigvee_{k=1}^{n}\bigl(M(i,k)\wedge M(j,k)\bigr).
\]

Hence
\[
\begin{array}{*{20}c}
   {M_Y (i,j) = 1} & { \Leftrightarrow } &
   {\exists \,k \in \{ 1, \ldots ,n\}
   \;(M(i,k) = 1 \wedge M(j,k) = 1)}  \\
   {} &  \Leftrightarrow  &
   {\exists \,x_k  \in X
   \;[(y_i ,x_k ) \in E \wedge (y_j ,x_k ) \in E].}
\end{array}
\]

Thus $M_Y$ is the matrix representation
of the co-source relation ${\cal Y}$.

\medskip
\noindent
\textbf{(2)}

From (1) and Definition~\ref{DF:BG}(3),
$M_Y$ represents a symmetric relation.
The closure $M_Y^*$ inherits symmetry,
while also extending reflexivity and transitivity.
Hence $B_Y=M_Y^*$
is the matrix representation
of an equivalence relation,
which induces the partition \eqref{EQ:YD}.

Next we prove
$\operatorname{Dom}(Y_p)\cap \operatorname{Dom}(Y_q)=\emptyset$ for $p\ne q$.

Suppose otherwise.
Then there exists
\[
\tilde X=
\operatorname{Dom}(Y_p)\cap \operatorname{Dom}(Y_q)
\ne\emptyset,
\quad p\ne q.
\]

Choose $\tilde x\in\tilde X$.
Then there exist
$y_p\in Y_p$ and $y_q\in Y_q$ such that
\[
(y_p,\tilde x)\in E\wedge (y_q,\tilde x)\in E.
\]
By the definition of the co-source relation,
$y_p$ and $y_q$ share a common source.
Thus they belong to the same equivalence class
under $B_Y$,
which implies $Y_p=Y_q$,
a contradiction.

\end{proof}

\medskip
\noindent
\textbf{Row compression of the equivalence-class matrix.}

Let $A$ be the matrix representation
of an equivalence relation $R$.
Then $(i,j)\in R$ implies
\[
A(i,:)=A(j,:).
\]

Hence one row from each equivalence class
may be selected as its representative.

The row vector $A(s,:)$
is simultaneously the Boolean indicator vector
of the equivalence class containing $s$:
\[
\{\,i\mid (s,i)\in R\,\}
=
\{\,i\mid A(s,i)=1\,\}.
\]

Thus the equivalence-class matrix
can be compressed so that
each row corresponds to exactly one equivalence class.

\begin{definition}\label{DF:BG-compression}

Let $B_Y=M_Y^*$ be the co-source equivalence matrix.
Let $R_Y$ be the compressed matrix of $B_Y$:
the rows of $R_Y$ are selected from $B_Y$,
taking exactly one row from each equivalence class.

$R_Y$ is called the
co-source equivalence-class indicator matrix.

\end{definition}

\begin{theorem}\label{THM:BG-compression}

Let $M$ be the mapping relation matrix
of the bipartite graph
$G=\langle Y\cup X,\,E\subset Y\times X,\,M\rangle$,
and let
$\bar Y=\{Y_1,\dots,Y_k\}$
be the co-source equivalence partition of $Y$.
Let $R_Y$ be its indicator matrix.
Then

\begin{enumerate}

\item
The $s$-th row of $R_Y$
is the Boolean indicator vector
of the co-source equivalence class
\[
Y_s=\{\,y_i\mid R_Y(s,i)=1\,\}.
\]

\item
\[
D_Y(s,:)=(R_Y\odot M)(s,:)
\]
is the Boolean indicator vector
of $\operatorname{Dom}(Y_s)$,
thus
\[
\operatorname{Dom}(Y_s)=\{\,x_j\mid D_Y(s,j)=1\,\}.
\]

\end{enumerate}

\end{theorem}

\begin{proof}

From Definition~\ref{DF:BG-compression},
$R_Y(s,:)$ corresponds to
the co-source equivalence class $Y_s$.
Hence
\[
Y_s=\{\,y_i\mid R_Y(s,i)=1\,\}.
\]

Moreover,
\[
(R_Y\odot M)(s,:)
=
\bigvee_{R_Y(s,i)=1} M(i,:)
=
\bigvee_{y_i\in Y_s} M(i,:).
\]
Since $M(i,:)$ is the Boolean indicator vector
of the source set
$\operatorname{Dom}(y_i)$,
their Boolean OR yields
the indicator vector
of the entire source set
$\operatorname{Dom}(Y_s)$,
namely $D_Y(s,:)$.

\end{proof}

\medskip
\noindent
\textbf{Obtaining maximal weakly connected subgraphs and diagonal blocks from co-source equivalence classes.}

Theorem~\ref{THM:BG-compression}
provides a method for partitioning maximal weakly connected subgraphs
without explicitly constructing
the adjacency matrix of the bipartite graph.

Each co-source equivalence class $Y_s$,
together with its source set $\operatorname{Dom}(Y_s)$,
forms a maximal weakly connected subgraph of the bipartite graph
\[
G_s=
\langle
Y_s\cup \operatorname{Dom}(Y_s),
E_s\subset Y_s\times \operatorname{Dom}(Y_s),
M^s
\rangle .
\]

Accordingly,
define permutations $P,Q$
to group together the rows $y_i$
and columns $x_j$
belonging to the same subgraph.
This yields a block-diagonal structure of $M$.

Note that the co-source equivalence classes
produce a partition of $Y$
and $\operatorname{Dom}(Y)$,
but may not cover all nodes in $X$.
Therefore,
all nodes $x_j$
that do not appear in
\[
\operatorname{Dom}(Y)
=
\bigcup_{s=1}^k
\operatorname{Dom}(Y_s)
\]
are merged into a single zero block $X_{k+1}$,
thereby producing
a dimensionally closed block-diagonal representation.

\subsection{Algorithm Description}

Let $M \in {0,1}^{m \times n}$.
The algorithm uses the following structural tables
to store intermediate computational results:
\[
\begin{array}{*{20}c}
   {Y < Index,SubGroup,NewIndex > } & , & {X < Index,SubGroup,NewIndex > }
 \end{array}
\]

The table $Y$ has $m$ rows,
and the table $X$ has $n$ rows.
Initially,
$Y.Index$ is assigned sequentially from $1$ to $m$,
and $X.Index$ is assigned sequentially from $1$ to $n$.
These represent the original indices of the components of $Y$ and $X$,
that is,
the original row and column indices of $M$.

During the computation,
each row of $Y$ and $X$
(that is,
each $y_i$ and $x_j$ under the original indexing)
is assigned a maximal weakly connected subgraph label
$SubGroup$.
The rows are then grouped according to $SubGroup$.

After grouping,
$Y.NewIndex$ is assigned sequentially from $1$ to $m$,
and $X.NewIndex$ is assigned sequentially from $1$ to $n$.

After the above processing is completed,
the following subtables of $Y$ and $X$:
\[
\begin{array}{*{20}c}
   {P =  < Y.NewIndex,Y.Index > } & {} & {Q =  < X.NewIndex,X.Index > }
 \end{array}
\]
give the required permutations.
Here,
$P$ is the row permutation of $M$,
and $Q^T$ is the column permutation of $M$.

\subsection{Basic Algorithm Procedure}

The following algorithm describes the procedure
based on the co-source relation (the $Y$-side construction).

\begin{enumerate}

\item Generate the $P$-mapping relation matrix $M$
from the parameter matrix $W$
(Definition~\ref{DF:BG}(2)).

\item Compute the co-source relation matrix of $Y$
(Theorem~\ref{THM:co-Dom}(1)):
\[
M_Y = M \odot M^T .
\]

\item Use the Kleene--Valiant algorithm
(Theorem~\ref{thm1}(2))
to compute the co-source equivalence relation matrix
(Theorem~\ref{THM:co-Dom}(2)):
\[
B_Y = M_Y^* .
\]

\item Based on $B_Y$,
generate the field $SubGroup$
in
$Y\langle Index,SubGroup,NewIndex\rangle$:
vertices belonging to the same equivalence class
are assigned the same label,
while different equivalence classes
receive different labels.
Let the number of equivalence classes be $k$
(i.e., the graph is partitioned into $k$ subgraphs).

\item Construct the compressed matrix
$R_Y\in\{0,1\}^{k\times m}$
from $Y.SubGroup$
(Definition~\ref{DF:BG-compression}):
\[
R_Y(i,j)=
\begin{cases}
1, & i=Y.SubGroup(j),\\
0, & \text{otherwise}.
\end{cases}
\]

\item Compute the source-set matrix
of each co-source equivalence class
(Theorem~\ref{THM:BG-compression}):
\[
D_Y = R_Y \odot M ,
\]
where $D_Y(s,:)$ is the Boolean indicator vector
of $\operatorname{Dom}(Y_s)$.

\item Generate the field $SubGroup$
of
$X\langle Index,SubGroup,NewIndex\rangle$
based on $D_Y$,
and handle the boundary case
of the “zero block”
($\operatorname{Dom}(Y_s)=\emptyset$):

\begin{enumerate}

\item[(7-1)]
If $D_Y(s,:)\neq \mathbf{0}$,
then for all $x_j$ satisfying $D_Y(s,j)=1$ set
\[
X.SubGroup(j)\leftarrow s .
\]

\item[(7-2)]
If $D_Y(s,:)=\mathbf{0}$,
then this equivalence class does not correspond
to a nonzero subgraph.
Merge the entire class into the zero block:
\[
Y.SubGroup(i)\leftarrow k+1,
\qquad Y.SubGroup(i)=s .
\]

\end{enumerate}

\item Complete the nodes in $X$
that are not covered by $\operatorname{Dom}(Y)$:
\[
X.SubGroup(j)\leftarrow k+1,
\qquad \forall\,X.SubGroup(j)=0 .
\]

\item Sort $Y$ and $X$
in ascending order of
$Y.SubGroup$ and $X.SubGroup$ respectively,
and assign $NewIndex$ sequentially:
\[
Y.NewIndex(i)=i \quad (i=1,\dots,m),
\]
\[
X.NewIndex(j)=j \quad (j=1,\dots,n).
\]

\end{enumerate}

After completing the above steps,
the permutations are obtained as
\[
P=\langle Y.NewIndex,\,Y.Index\rangle,
\qquad
Q=\langle X.NewIndex,\,X.Index\rangle .
\]
Here $P$ is the row permutation of $M$,
and $Q^T$ is the column permutation of $M$.

\subsection{Algorithm Description (Procedural Pseudocode)}\label{SEC:AB}

The overall structure of the algorithm is a sequential procedure.
Except for a single loop required to generate labels
from the matrix computation results
and write them into the node attribute tables,
no nested control structure is involved.
For readers from different backgrounds,
the algorithm is described below using procedural pseudocode
rather than a flowchart.

\begin{enumerate}

\item \textbf{Input:}
parameter matrix $W\in\mathbb{R}^{m\times n}$.

\item \textbf{Initialize matrices and node attribute tables:}

\begin{enumerate}
\item[(2-1)]
$M^{m\times n}\leftarrow 0$
($P$-mapping matrix, Boolean);

\item[(2-2)]
$M_Y^{m\times m}\leftarrow 0$
(co-source relation matrix);

\item[(2-3)]
$B_Y^{m\times m}\leftarrow 0$
(co-source equivalence relation matrix);

\item[(2-4)]
initialize the node attribute tables
\[
Y\langle Index,SubGroup,NewIndex\rangle,\quad
X\langle Index,SubGroup,NewIndex\rangle
\]
and set
\[
Y.Index(i)=i,\;
X.Index(j)=j,\;
Y.SubGroup(i)=0,\;
X.SubGroup(j)=0,
\]

where $i=1,\dots,m$ and $j=1,\dots,n$.

\end{enumerate}

\item \textbf{Construct the $P$-mapping matrix $M$}
(Definition~\ref{DF:BG}(2)):
\[
m_{ij}=
\begin{cases}
1, & P(w_{ij}),\\
0, & \neg P(w_{ij}),
\end{cases}
\qquad
i=1,\dots,m,\ j=1,\dots,n.
\]

\item \textbf{Compute the co-source relation matrix}
(Theorem~\ref{THM:co-Dom}(1)):
\[
M_Y \leftarrow M\odot M^T .
\]

\item \textbf{Compute the co-source equivalence relation matrix}
(Theorem~\ref{THM:co-Dom}(2)):

\begin{enumerate}
\item[(5-1)]
$B_Y \leftarrow M_Y \vee I$;
\item[(5-2)]
For $t=1$ to $\lceil \log_2(m-1)\rceil$ Do
\[
B_Y \leftarrow B_Y \odot B_Y .
\]

\end{enumerate}

\item \textbf{Determine the co-source equivalence class labels of $Y$:}

\begin{enumerate}

\item[(6-1)]
$SGIndex\leftarrow 1$;

\item[(6-2)]
While there exists $i$ such that $Y.SubGroup(i)=0$ Do
\[
Y.SubGroup \leftarrow
Y.SubGroup + SGIndex\cdot B_Y(i,:)^{T};
\]
(here $B_Y(i,:)$ is the Boolean indicator vector
of the co-source equivalence class containing node $y_i$.
Multiplying by the current subgraph label $SGIndex$
assigns the same subgraph label
to all nodes in this equivalence class simultaneously.)
\[
SGIndex \leftarrow SGIndex + 1 .
\]
\item[(6-3)]
Number of equivalence classes
$k\leftarrow SGIndex-1$.

\end{enumerate}

\item \textbf{Initialize the compression matrix and source-set matrix:}

\begin{enumerate}

\item[(7-1)]
$R_Y^{k\times m}\leftarrow 0$
(co-source equivalence-class matrix);

\item[(7-2)]
$D_Y^{k\times n}\leftarrow 0$
(source-set matrix).

\end{enumerate}

\item \textbf{Generate the compressed co-source equivalence matrix $R_Y$:}

For $p=1$ to $m$ Do
\[
s\leftarrow Y.SubGroup(p),\qquad
R_Y(s,p)\leftarrow 1 .
\]

\item \textbf{Compute the source-set matrix of the co-source equivalence classes}
(Theorem~\ref{THM:BG-compression}):
\[
D_Y \leftarrow R_Y \odot M .
\]

\item \textbf{Determine $X.SubGroup$ based on $D_Y$,
and handle equivalence classes without sources:}

For $s=1$ to $k$ Do
\begin{itemize}
\item
If $D_Y(s:)\neq \mathbf{0}$, then
\[
X.SubGroup \leftarrow
X.SubGroup + s\cdot D_Y(s,:)^{T};
\]
\item
otherwise,
set nodes satisfying $Y.SubGroup=s$ to
\[
Y.SubGroup \leftarrow k+1 .
\]
\end{itemize}

\item \textbf{Handle $X$ nodes not contained in $\operatorname{Dom}(Y)$:}

For all $X.SubGroup(j)=0$,
set
\[
X.SubGroup(j)\leftarrow k+1 .
\]

\item \textbf{Generate the permutations:}

Sort $Y$ and $X$
in ascending order of $SubGroup$ respectively,
and assign sequentially
\[
Y.NewIndex(i)=i,\quad i=1,\dots,m;
\]
\[
X.NewIndex(j)=j,\quad j=1,\dots,n .
\]

\item \textbf{Output:}
\[
P=\langle Y.NewIndex,Y.Index\rangle,\quad
Q=\langle X.NewIndex,X.Index\rangle .
\]

\end{enumerate}

\subsection{Algorithm Example}

This subsection happens to provide a very good example
for understanding the concept of permutation
and its role in vector and matrix operations.
We therefore use this case study
to further clarify the concept of permutation
and to introduce some practical details of its application.

For the basic notion of permutation,
see Section~\ref{permutation} of this paper.
For more detailed discussions on permutation,
one may consult the literature on permutation groups
\cite{armstrong1988groups, Seress2003Permutation, Sims1971Computation}.
However,
this paper does not require deep knowledge of permutation group theory.
Only the following points are needed.

1. A permutation is a bijection
defined on the finite integer set
$I_n=\{1,\dots,n\}$:
\[
I_n \xrightarrow{P} I_n .
\]
Since it is a mapping on a finite set,
the most common way to represent a permutation
is by a value table:
\begin{equation}\label{EQ:permutation}
\begin{array}{*{20}c}
   {P = \left( {\begin{array}{*{20}c}
   1 & {s_1 }  \\
   2 & {s_2 }  \\
    \vdots  &  \vdots   \\
   n & {s_n }  \\
 \end{array} } \right)} &
 {P^{ - 1}  = \left( {\begin{array}{*{20}c}
   {s_1 } & {\text{1}}  \\
   {s_2 } & 2  \\
    \vdots  &  \vdots   \\
   {s_n } & n  \\
 \end{array} } \right)} &
 {P^T  = \left( {\begin{array}{*{20}c}
   1 & 2 &  \cdots  & n  \\
   {s_1 } & {s_2 } &  \cdots  & {s_n }  \\
 \end{array} } \right)}
\end{array}
\end{equation}

In this representation,
the left column of permutation $P$
contains the values of the target variable,
and the right column contains the values of the source variable.

The greatest advantage of representing a permutation
by a value table
is that it is computation-oriented,
as will become clear in the following example.
A basic property of permutations is that
they are independent of the order in which the rows are arranged.
Different row orders represent the same permutation.
For this reason,
the first column of the value table
is usually arranged in ascending order.

Historically,
the most common use of permutations
was to reorder the rows and columns
of vectors or matrices (determinants).
For example,
when the above permutation $P$
acts on a column vector $X$
and a matrix $M$,
the meanings of $PX$ and $PM$
are respectively:
\[
\begin{array}{*{20}c}
   {(PX)(i) = X(s_i )} & , & {(PM)(i,:) = M(s_i ,:).}
 \end{array}
\]

2. A permutation forms an algebraic system
(the permutation group).

(2-1)
Two permutations of the same order
can be composed as functions.
For a permutation $R$,
we use $R(i)$ to denote
the value in the right column
corresponding to the left-column value $i$.
Let $P,Q$ be permutations of order $n$.
Then the composition of $P$ and $Q$ is
\[
\begin{array}{*{20}c}
   {i \to P(i) = j \to Q(j)} &  \Rightarrow  & {(PQ)(i) = Q(P(i))}
 \end{array}
\]

Hence,
\[
PQ = \left( {\begin{array}{*{20}c}
   1 & {Q(s_1 )}  \\
   2 & {Q(s_2 )}  \\
    \vdots  &  \vdots   \\
   n & {Q(s_n )}  \\
 \end{array} } \right)
=
\left( {\begin{array}{*{20}c}
   1 & {Q(P(1))}  \\
   2 & {Q(P(2))}  \\
    \vdots  &  \vdots   \\
   n & {Q(P(n))}  \\
 \end{array} } \right)
\]

(2-2)
Since a permutation is a bijection,
its inverse exists.
In the value-table representation,
$P^{-1}$ is obtained
by exchanging the two columns of $P$
in (\ref{EQ:permutation}).

3. In (\ref{EQ:permutation}),
the horizontal arrangement of $P$
is exactly the transpose $P^T$ of $P$.
A vertically arranged value table,
such as $P$ in (\ref{EQ:permutation}),
is usually used to define a row permutation;
a horizontally arranged value table,
such as $P^T$ in (\ref{EQ:permutation}),
is usually used to define a column permutation.

It should be emphasized that
this is not a special convention of the present paper.
In the era of hand computation of determinants
and matrix permutations,
such conventions were already standard.
At that time,
a permutation never meant a “permutation matrix,”
because the difference in computational workload
was enormous.
Even in the later era,
when permutation matrices were used
as matrix representations of permutations
(mainly for theoretical convenience),
mathematicians still preserved
the formal consistency between the two notations
as much as possible.
Under this convention,
the symbol
\[
PMP^T
\]
has the same meaning
whether $P$ is viewed as a permutation
or as a permutation matrix.

We now present one execution example of the algorithm
and analyze its result.

\paragraph{(1) Original matrix.}
\begin{equation}\label{EQ:Ex-B-M}
M=
\left(
\begin{array}{cccc|ccc|ccc|cc}
1&0&1&1&0&0&0&0&0&0&0&0\\
0&1&1&0&0&0&0&0&0&0&0&0\\
1&1&0&1&0&0&0&0&0&0&0&0\\ \hline
0&0&0&0&1&0&1&0&0&0&0&0\\
0&0&0&0&1&1&0&0&0&0&0&0\\
0&0&0&0&1&1&1&0&0&0&0&0\\
0&0&0&0&1&1&0&0&0&0&0&0\\ \hline
0&0&0&0&0&0&0&1&1&0&0&0\\
0&0&0&0&0&0&0&1&1&0&0&0\\
0&0&0&0&0&0&0&0&1&1&0&0\\ \hline
0&0&0&0&0&0&0&0&0&0&0&0\\
0&0&0&0&0&0&0&0&0&0&0&0
\end{array}
\right).
\end{equation}

\paragraph{(2) Scrambling by random permutations.}

Take a pair of random permutations
(value-table semantics:
\emph{new $\leftarrow$ old})
\[
P_1  = \left( {\begin{array}{c|cccccccccccc}
   {{\text{new}}} & 1 & 2 & 3 & 4 & 5 & 6 & 7 & 8 & 9 & {10} & {11} & {12}  \\
   {{\text{old}}} & 8 & {12} & 4 & {11} & 9 & 5 & {10} & 2 & 1 & 7 & 3 & 6  \\
 \end{array} } \right)^T
\]
\[
Q_1  = \left( {\begin{array}{c|cccccccccccc}
   {{\text{new}}} & 1 & 2 & 3 & 4 & 5 & 6 & 7 & 8 & 9 & {10} & {11} & {12}  \\
   {{\text{old}}} & 8 & 5 & 3 & 6 & {12} & {10} & {11} & 7 & 2 & 1 & 4 & 9  \\
 \end{array} } \right)^T
\]

The permutation $P_1$
moves row $8$ of $M$ to row $1$,
row $12$ to row $2$, and so on.
After the row permutation,
the column permutation $Q_1^T$ is applied,
moving column $8$ to column $1$,
column $5$ to column $2$, and so on.
Reordering the rows and columns of $M$
gives:
\[
M^{\prime}=P_1 M Q^T_1
=\left(\begin{array}{cccccccccccc}
1&0&0&0&0&0&0&0&0&0&0&1\\
0&0&0&0&0&0&0&0&0&0&0&0\\
0&1&0&0&0&0&0&1&0&0&0&0\\
0&0&0&0&0&0&0&0&0&0&0&0\\
1&0&0&0&0&0&0&0&0&0&0&1\\
0&1&0&1&0&0&0&0&0&0&0&0\\
0&0&0&0&0&1&0&0&0&0&0&1\\
0&0&1&0&0&0&0&0&1&0&0&0\\
0&0&1&0&0&0&0&0&0&1&1&0\\
0&1&0&1&0&0&0&0&0&0&0&0\\
0&0&0&0&0&0&0&0&1&1&1&0\\
0&1&0&1&0&0&0&1&0&0&0&0
\end{array}\right)
\]

This matrix is taken
as the input matrix of our algorithm.

\paragraph{(3) Run the algorithm and output the $Y/X$ tables.}

Apply to $M'$
the permutation-construction algorithm
for parameter-matrix structural decomposition
of \textbf{neural networks without feedback}
described in Section~\ref{SEC:AB}.
The algorithm outputs the following two structural tables:
\[
Y=\left(\begin{array}{c|c|c}
\text{Index}&\text{SubGroup}&\text{NewIndex}\\ \hline
1&1&1\\
5&1&2\\
7&1&3\\
3&3&4\\
6&3&5\\
10&3&6\\
12&3&7\\
8&5&8\\
9&5&9\\
11&5&10\\
2&6&11\\
4&6&12
\end{array}\right),
\qquad
X=\left(\begin{array}{c|c|c}
\text{Index}&\text{SubGroup}&\text{NewIndex}\\ \hline
1&1&1\\
6&1&2\\
12&1&3\\
2&3&4\\
4&3&5\\
8&3&6\\
3&5&7\\
9&5&8\\
10&5&9\\
11&5&10\\
5&6&11\\
7&6&12
\end{array}\right).
\]

\paragraph{(4) Construct permutations from the $Y/X$ tables.}

According to the algorithm definition, take
\[
P_2=\langle Y.\text{NewIndex},\,Y.\text{Index}\rangle
=
\left(
\begin{array}{c|cccccccccccc}
\text{new} & 1&2&3&4&5&6&7&8&9&10&11&12\\ 
\text{old} & 1&5&7&3&6&10&12&8&9&11&2&4
\end{array}
\right)^T,
\]
\[
Q_2=\langle X.\text{NewIndex},\,X.\text{Index}\rangle
=
\left(
\begin{array}{c|cccccccccccc}
\text{new} & 1&2&3&4&5&6&7&8&9&10&11&12\\ 
\text{old} & 1&6&12&2&4&8&3&9&10&11&5&7
\end{array}
\right)^T.
\]

\paragraph{(5) Reordering result of the algorithm.}

Applying the row and column permutations
$P_2,Q_2$ to $M'$
gives $M^{\prime \prime}$.
\[
M^{\prime\prime}=P_2 M^{\prime} Q^T_2
=
\left(
\begin{array}{ccc|ccc|cccc|cc}
1&0&1&0&0&0&0&0&0&0&0&0\\
1&0&1&0&0&0&0&0&0&0&0&0\\
0&1&1&0&0&0&0&0&0&0&0&0\\ \hline
0&0&0&1&0&1&0&0&0&0&0&0\\
0&0&0&1&1&0&0&0&0&0&0&0\\
0&0&0&1&1&0&0&0&0&0&0&0\\
0&0&0&1&1&1&0&0&0&0&0&0\\ \hline
0&0&0&0&0&0&1&1&0&0&0&0\\
0&0&0&0&0&0&1&0&1&1&0&0\\
0&0&0&0&0&0&0&1&1&1&0&0\\ \hline
0&0&0&0&0&0&0&0&0&0&0&0\\
0&0&0&0&0&0&0&0&0&0&0&0
\end{array}
\right).
\]

$M^{\prime \prime}$ achieves the goal
of structurally decomposing $M^{\prime}$,
but it is not exactly identical to $M$.
A preliminary observation is that
the first and third blocks of $M$
have exchanged positions in $M^{\prime \prime}$,
and the internal order within each block
is also not exactly the same.
This can be explained by the fact that
the orders of $y_i$ and $x_j$
within each block
are different in $M^{\prime \prime}$ and in $M$.

\paragraph{(6) Composite permutations and closed-loop verification.}

Is there a way to verify that
$M$ and $M^{\prime \prime}$
correspond to the same matrix
under two different block arrangements?
That is,
are $M$ and $M^{\prime \prime}$
different only in the ordering of subgraphs
and the ordering of nodes within each subgraph?
Let us examine how $M^{\prime \prime}$ is obtained.
\[
M'' = P_2 M'Q_2^T
= P_2 \left( {P_1 MQ_1^T } \right)Q_2^T
= \left( {P_2 P_1 } \right)M\left( {Q_1^T Q_2^T } \right)
\]
Therefore,
\[
M = \left( {P_2 P_1 } \right)^{ - 1} M''\left( {Q_1^T Q_2^T } \right)^{ - 1}
= \left( {P_2 P_1 } \right)^{ - 1} M''\left( {\left( {Q_2 Q_1 } \right)^T } \right)^{ - 1}
\]

Hence,
if the above transformation maps $M''$ back to $M$,
then the answer to the previous question is obtained.

According to the value-table composition rule
\[
(PQ)(i)=Q(P(i)),
\]
we compute
\[
P_2 P_1  = \left( {\begin{array}{c|cccccccccccc}
   {{\text{new}}} & 1 & 2 & 3 & 4 & 5 & 6 & 7 & 8 & 9 & {10} & {11} & {12}  \\
   {{\text{old}}} & 8 & 9 & {10} & 4 & 5 & 7 & 6 & 2 & 1 & 3 & {12} & {11}  \\
 \end{array} } \right)^T
\]
\[
Q_2 Q_1  = \left( {\begin{array}{c|cccccccccccc}
   {{\text{new}}} & 1 & 2 & 3 & 4 & 5 & 6 & 7 & 8 & 9 & {10} & {11} & {12}  \\
   {{\text{old}}} & 8 & {10} & 9 & 5 & 6 & 7 & 3 & 2 & 1 & 4 & {12} & {11}  \\
 \end{array} } \right)^T
\]

Exchange the two columns of
$P_2 P_1$ and $Q_2 Q_1$,
and sort them by the values in the first column
(note that $P_2 P_1$ and $Q_2 Q_1$
are vertically arranged value tables):
\[
(P_2 P_1 )^{ - 1}  = \left( {\begin{array}{c|cccccccccccc}
   {{\text{new}}} & 1 & 2 & 3 & 4 & 5 & 6 & 7 & 8 & 9 & {10} & {11} & {12}  \\
   {{\text{old}}} & 9 & 8 & {10} & 4 & 5 & 7 & 6 & 1 & 2 & 3 & {12} & {11}  \\
 \end{array} } \right)^T
\]
\[
(Q_2 Q_1 )^{ - 1}  = \left( {\begin{array}{*{20}c}
   {{\text{new}}} & 1 & 2 & 3 & 4 & 5 & 6 & 7 & 8 & 9 & {10} & {11} & {12}  \\
   {{\text{old}}} & 9 & 8 & 7 & {10} & 4 & 5 & 6 & 1 & 3 & 2 & {12} & {11}  \\
 \end{array} } \right)^T
\]

Direct calculation verifies that
\[
(P_2P_1)^{-1} M^{\prime\prime} ((Q_2Q_1)^T)^{-1}
=
\left(
\begin{array}{cccc|ccc|ccc|cc}
1&0&1&1&0&0&0&0&0&0&0&0\\
0&1&1&0&0&0&0&0&0&0&0&0\\
1&1&0&1&0&0&0&0&0&0&0&0\\ \hline
0&0&0&0&1&0&1&0&0&0&0&0\\
0&0&0&0&1&1&0&0&0&0&0&0\\
0&0&0&0&1&1&1&0&0&0&0&0\\
0&0&0&0&1&1&0&0&0&0&0&0\\ \hline
0&0&0&0&0&0&0&1&1&0&0&0\\
0&0&0&0&0&0&0&1&1&0&0&0\\
0&0&0&0&0&0&0&0&1&1&0&0\\ \hline
0&0&0&0&0&0&0&0&0&0&0&0\\
0&0&0&0&0&0&0&0&0&0&0&0
\end{array}
\right)
= M.
\]

\section{Structural Decomposition of General Directed Networks}

This section corresponds to the structural decomposition
of parameter matrices in recurrent / feedback neural networks.

\subsection{Method Principle}

In artificial neural networks with feedback structure,
the parameter matrix is usually square,
and its $i$-th row and $i$-th column
correspond respectively
to the incoming-edge and outgoing-edge relations
of the same neuron node.
Such networks correspond,
in graph-structural terms,
to a general directed graph,
in which cycles and strongly connected components are allowed.

The parameter matrix
$W^{n\times n}$
may be regarded as the adjacency-matrix representation
of a directed graph
$G=\langle V,E\rangle$,
where
\[
W(i,j)\neq 0 \iff (j\to i)\in E.
\]

After training is completed,
the parameter matrix typically exhibits
a two-level structure at the structural level.

\paragraph{First level: weakly connected subgraphs.}

A directed graph $G$
can be decomposed into a number of weakly connected subgraphs.
In the parameter matrix,
these subgraphs correspond to a collection of mutually uncoupled diagonal blocks.
Through an appropriate permutation,
$W$ can be reordered into the block-diagonal form
\[
PWP^{-1}
=
\mathrm{diag}(W_1,W_2,\dots,W_k),
\]
where each $W_i$
corresponds to one weakly connected subgraph,
and no dependency exists between different subgraphs.
This level characterizes
the partition of the system
into mutually independent subsystems.

\paragraph{Second level: strongly connected components and sequential dependencies.}

Within each weakly connected subgraph,
its strongly connected components (SCCs)
can be identified.
The SCCs form a condensation graph,
which is necessarily a directed acyclic graph.
Therefore,
within each diagonal block,
an appropriate permutation
can transform $W_i$
into a block-triangular
(Frobenius) normal form
\cite{horn2013matrix},
whose diagonal subblocks correspond
to irreducible strongly connected structures,
while the off-diagonal blocks encode
the sequential dependency relations
among those diagonal subblocks.

In what follows,
we give an algorithm
based entirely on matrix operations
to uniformly realize
the identification of the above two-level structure
and the construction of the corresponding permutations.

\begin{definition}\label{def5}

A Boolean matrix
$M\in\{0,1\}^{n\times n}$
is called the $P$-adjacency matrix of $W$
(abbreviated as the $P$-adjacency matrix),
defined by
\[
M(i,j)=
\begin{cases}
1, & P(W(i,j)),\\
0, & \neg P(W(i,j)),
\end{cases}
\qquad i,j=1,\dots,n.
\]
Here $P(x)$ means that $x$ has property $P$,
and $\neg P(x)$ means that $x$ does not have property $P$.

\end{definition}

\begin{definition}\label{defn2}

Let $G=\langle V,E\rangle$
be a simple directed graph,
and let $A$ be its adjacency matrix.
Let
$B=A^* \wedge (A^*)^T$
be the matrix representation
of the mutual-reachability relation on $V$.
Let
$\bar V = \{V_1,V_2,\dots,V_s\}$
be the collection of mutual-reachability equivalence classes.
Each equivalence class $V_p$
may be viewed as an index set:
$i\in V_p$
corresponds to the vertex of $G$
whose index is $i$.

\begin{enumerate}
\item
The subgraph
$G_p=\langle V_p,\ E_p= (V_p\times V_p)\cap E\rangle$
is called a maximal strongly connected subgraph of $G$,
or equivalently,
a strongly connected component (SCC) of $G$.

\item
$\bar G=\langle \bar V,\bar E\rangle$
is called the condensation graph of $G$.
The vertices of the condensation graph
are the SCCs of $G$.
For $V_p\neq V_q$,
\[
(V_p ,V_q ) \in \bar E
\Leftrightarrow
\exists \,i \in V_p \exists \,j \in V_q
[(v_i ,v_j ) \in E].
\]

\item
Let $s$ be the order of the condensation graph $\bar G$
(that is, the number of SCCs),
and let $n$ be the order of $G$.
A matrix
$R\in\{0,1\}^{s\times n}$
is called the SCC condensation compression matrix:
its rows are selected from the mutual-reachability matrix $B$,
taking exactly one row from each equivalence class.

\end{enumerate}
\end{definition}

\begin{theorem}\label{thm4}

Let $G=\langle V,E\rangle$
be a simple directed graph,
let $A$ be its adjacency matrix,
and let
$\bar G=\langle \bar V,\bar E\rangle$
be its condensation graph.

\begin{enumerate}
\item
The condensation graph $\bar G$
is acyclic (a DAG),
with the following corollaries:
\begin{enumerate}
\item[(1-1)]
$\bar G$ contains at least one vertex of in-degree zero;
\item[(1-2)]
after removing all vertices of in-degree zero,
the remaining graph is still a layered acyclic graph.
\end{enumerate}

\item
Let $R$ be the condensation compression matrix of $G$,
and let $I_s$ be the $s\times s$ Boolean identity matrix.
If $\bar A$ denotes the adjacency matrix
of the condensation graph,
then
\[
\bar A = \left(R \odot A \odot R^T\right)\wedge \neg I_s .
\]
\end{enumerate}
\end{theorem}

\begin{proof} \

(1)
By definition,
$\bar G$ does not contain edges of the form
$(V_p, V_p) \in \bar E$.
If a cycle existed,
it would necessarily consist of different mutual-reachability equivalence classes.
But then those equivalence classes would merge into a single mutual-reachability class,
that is,
they would in fact be the same equivalence class,
which contradicts the assumption.

(1-1)
Starting from $V_1$,
trace backward a longest path
without repeated vertices:
\[
  V_k  \to  \cdots  \to V_1 .
\]
The in-degree of $V_k$ must be zero.
Suppose instead that there exists
$V_s \to V_k$.
If
$V_s \not \in \{V_1, \dots ,V_k\}$,
then the above path is not the longest path
without repeated vertices.
If
$V_s \in \{V_1, \dots ,V_k\}$,
then a cycle exists,
contradicting (1).

(1-2)
Since removing vertices of in-degree zero
does not introduce any new cycle,
the conclusion is immediate.

(2)
The $p$-th row of the condensation compression matrix $R$
is taken from the mutual-reachability matrix $B$
and corresponds to the strongly connected component $V_p$.
Hence
\[
R(p,i)=1 \iff i\in V_p.
\]

Consider the Boolean product
$(R\odot A\odot R^T)(p,q)$.
First,
\[
(R\odot A)(p,j)
= \bigvee_{i=1}^n (R(p,i)\wedge A(i,j))
= \bigvee_{i\in V_p} A(i,j).
\]
Therefore,
\[
(R\odot A)(p,j)=1
\iff
\exists\, i\in V_p \ \text{s.t.}\ A(i,j)=1.
\]

Furthermore,
\[
(R\odot A\odot R^T)(p,q)
= \bigvee_{j=1}^n \bigl((R\odot A)(p,j)\wedge R^T(j,q)\bigr)
= \bigvee_{j\in V_q} (R\odot A)(p,j).
\]
Hence
\[
(R \odot A \odot R^T )(p,q)  = 1
\Leftrightarrow
\exists \,j \in V_q \;\exists \,i \in V_p [A(i,j)  = 1].
\]

By the definition of the adjacency matrix,
\[
A(i,j)=1 \iff (v_i, v_j)\in E,
\]
and the above condition is precisely
the criterion for the existence of an edge
$(V_p,V_q)\in\bar E$
in the condensation graph.

Therefore,
$R\odot A\odot R^T$
gives the adjacency relation
of the condensation graph.
Since the condensation graph does not retain self-loops,
taking the elementwise conjunction with $\neg I_s$
yields the adjacency-matrix representation
of the condensation graph.

\end{proof}

\subsection{Algorithm Description}\label{subsec:alg-explain}

The algorithm progressively computes
the adjacency matrix $M$ of a directed graph
and the adjacency matrix $M_C$ of its condensation graph,
thereby determining,
in the structural sense,
several key attributes of each node,
including:
the strongly connected component (SCC) to which the node belongs,
the hierarchical level of that SCC within the condensation graph,
the weakly connected component to which the node belongs,
and whether the node becomes an isolated point
after structural annealing is completed.

The algorithm records the results
of each stage of computation
by maintaining a node attribute table
\[
V \langle
Index,\,
S_{TAG},\,
G_{TAG},\,
L_{TAG},\,
I_{TAG},\,
NewIndex
\rangle^{n}
\]
where the meaning of each field is as follows:
\begin{itemize}
  \item $Index$: the original index of the node;
  \item $S_{TAG}$: the index of the strongly connected component (SCC) to which the node belongs;
  \item $G_{TAG}$: the index of the weakly connected component to which the node belongs;
  \item $L_{TAG}$: the layer index of the SCC containing the node in the condensation graph;
  \item $I_{TAG}$: the structural isolated-point indicator
  (a node takes value $1$ if, after annealing, it has neither incoming nor outgoing edges; otherwise it takes value $0$);
  \item $NewIndex$: the new node index generated according to the structural classification rules.
\end{itemize}

After all node attributes have been computed,
the algorithm sorts the nodes
according to the following priority:
first by the weakly connected component label $G_{TAG}$,
then by the isolated-point indicator $I_{TAG}$,
then by the layer index $L_{TAG}$,
and finally by the strongly connected component label $S_{TAG}$.
This generates the new node numbering $NewIndex$
and yields the required node permutation relation
\[
P =
<NewIndex,Index>.
\]

\subsection{Basic Algorithm Procedure}\label{subsubsec:principle-steps}

The following gives a matrix-based step-by-step description
for obtaining the node permutation $P$
from the parameter matrix $W$.

\begin{enumerate}

\item \textbf{Construct the $P$-adjacency matrix from the parameter matrix.}

From the parameter matrix $W$
(before or after annealing),
construct its $P$-adjacency matrix $M$
(Definition~\ref{def5}):
\[
m_{ij}=
\begin{cases}
1, & P(w_{ij}),\\
0, & \neg P(w_{ij}),
\end{cases}
\qquad i,j=1,\dots,n.
\]

Here $P(x)$ means that $x$ has property $P$,
and $\neg P(x)$ means that $x$ does not have property $P$.

\item \textbf{Compute the SCC labels $S_{TAG}$.}

Obtain the strongly connected component (SCC) label $S_{TAG}$
of each node by computing on $M$:

\begin{enumerate}

\item[(2-1)]
Use the Kleene--Valiant recursion
to compute the transitive $*$-closure $M^*$ of $M$
(Theorem~\ref{thm1}(2)).

\item[(2-2)]
Obtain the mutual-reachability relation of nodes
from the matrix
\[
B = M^* \wedge (M^*)^T
\]
(Theorem~\ref{thm1}(3)).

\item[(2-3)]
According to the partition into equivalence classes
induced by $B$,
determine the SCC label $S_{TAG}$
for each node in the node attribute table $V$
(Theorem~\ref{thm1}(3)).

\end{enumerate}

\item \textbf{Construct the condensation-graph adjacency matrix $M_C$.}

\begin{enumerate}

\item[(3-1)]
From the correspondence between node indices
and SCC labels in $V$,
construct the condensation compression matrix
$R_C^{k\times n}$
(Definition~\ref{defn2}(3)).

\item[(3-2)]
Use the condensation-graph matrix formula
\[
M_C = (R_C \odot M \odot R_C^T)\wedge \neg I_k
\]
to obtain the condensation-graph adjacency matrix $M_C$
(Theorem~\ref{thm4}(2)).

\end{enumerate}

\item \textbf{Compute the SCC layer labels $L_{TAG}$.}

By Theorem~\ref{thm4}(1),
the condensation graph is a DAG.
Hence the layer labels can be determined
by repeatedly peeling off
the set of nodes with zero in-degree level by level.

Let
$\mathbf{1}=(1,1,\dots,1)^T$
be the all-ones Boolean column vector,
and let
$S_{comp}$
denote the set of SCCs already processed
(initially $S_{comp}=\emptyset$).

Through
\begin{equation}\label{eq:SCCL}
Y = (M_C \wedge \text{Mask}(S_{comp})) \odot \mathbf{1}
\end{equation}
we obtain the in-degree-existence vector $Y$
for each SCC in the currently unprocessed part.

Here $\odot$
denotes Boolean matrix--vector multiplication
(Definition~\ref{defn1}(1)).

Note that
\[
Y(i)=\bigvee_{j=1}^{k}\left(M_C(i,j)\right),
\]
so when $Y(i)=0$,
the $i$-th SCC has zero in-degree
in the currently unprocessed condensation graph
(since this paper adopts the convention $j\to i$
for the adjacency matrix,
the in-degree corresponds to the Boolean OR of a row).
It can therefore be determined
that this SCC belongs to the current layer
(Theorem~\ref{thm4}(1-1)).

Add the SCCs of the current layer to $S_{comp}$,
and in the next iteration
use the mask matrix $\text{Mask}(S_{comp})$
to “mask out” the rows and columns
corresponding to the already processed SCCs.
Repeating this process yields the layer labels $L_{TAG}$
for all SCCs
(Theorem~\ref{thm4}(1-2)).

In actual implementation,
there is no need to physically delete matrix rows or columns;
it suffices to zero them out by masking.

\item \textbf{Compute the weakly connected component labels $G_{TAG}$.}

The weakly connected component labels
are computed on the condensation graph
rather than on the original graph,
because each SCC belongs to exactly one weakly connected component,
and the order of the condensation graph,
namely $k$ (the number of SCCs),
is usually much smaller than $n$.

\begin{enumerate}

\item[(5-1)]
Construct the underlying graph
of the condensation graph
(Definition~\ref{defn1}(3)):
\[
B_C= M_C \vee M_C^T .
\]
\item[(5-2)]
Use the Kleene--Valiant recursion
to compute the transitive $*$-closure
$B_C^*$
of $B_C$
(Theorem~\ref{thm1}(2)).
\item[(5-3)]
Use the equivalence-class partition
induced by $B_C^*$
to determine the weakly connected component label
$G_{TAG}$
for each SCC
(Theorem~\ref{thm1}(4)).

\end{enumerate}

\item \textbf{Structural isolated-point indicator $I_{TAG}$.}

Count the number of nodes contained
in each weakly connected component.
If a weakly connected component contains only one node,
then set the corresponding node's
structural isolated-point indicator to
$I_{TAG}=1$;
otherwise set
$I_{TAG}=0$.

\item \textbf{Sort and generate the new index $NewIndex$.}

Sort the node attribute table
\[
V\langle Index,S_{TAG},G_{TAG},L_{TAG},I_{TAG},NewIndex\rangle^{n}
\]
according to the following priority:
$G_{TAG}$, $I_{TAG}$, $L_{TAG}$, $S_{TAG}$.

Then assign consecutive values from top to bottom
to generate the new index $NewIndex$.

\item \textbf{Output the permutation.}

The algorithm outputs the node attribute table.
Among it, the subtable
\[
P=\langle NewIndex,Index\rangle
\]
is exactly the required node permutation.

\end{enumerate}

\subsection{Algorithm Description: Procedural Pseudocode}\label{subsec:pseudocode}

Structurally, this algorithm is relatively simple.
Except for a single loop required to generate labels
from matrix-computation results
and write them into the node attribute table,
the overall procedure is sequential.
For ease of reading by readers from different backgrounds,
a procedural description is adopted here
instead of a pseudocode flowchart.

\paragraph{Main procedure.}

\begin{enumerate}

\item Input the matrix order $n$.

\item Construct global matrices/tables:
\begin{enumerate}
\item[(2-1)] $W^{n\times n}$: input parameter matrix (numeric);
\item[(2-2)] $M^{n\times n}=0$: $P$-adjacency matrix (Boolean);
\item[(2-3)] $M^{*^{n\times n}}=0$: transitive $*$-closure of $M$ (Boolean);
\item[(2-4)] $B^{n\times n}=0$: mutual-reachability matrix
$B=M^*\wedge (M^*)^T$ (Boolean);
\item[(2-5)]
node attribute table
$V\langle Index,S_{TAG},G_{TAG},L_{TAG},I_{TAG},NewIndex\rangle^n$,
and initialize
\[
Index(i)=i,\qquad i=1,\dots,n.
\]
\item[(2-6)] condensation-graph order $k=0$ (output by a subprocedure);
\item[(2-7)]
condensation-graph adjacency-matrix cache
$C_M^{n\times n}=0$
(Boolean; the final output takes the upper-left $k\times k$ block).
\end{enumerate}

\item Input $W$.

\item Construct the $P$-adjacency matrix $M$ from $W$
(Definition~\ref{def5}):
\[
m_{ij}=
\begin{cases}
1, & P(w_{ij}),\\
0, & \neg P(w_{ij}),
\end{cases}
\qquad i,j=1,\dots,n.
\]

\item Compute $M^*$
(Kleene--Valiant recursion, Theorem~\ref{thm1}(2)):
$$M^* = M \vee I;$$
For $t=1$ to $\left\lceil \log_2 (n-1)\right\rceil$ Do
$$M^* \Leftarrow M^* \odot M^*.$$

\item Compute the mutual-reachability matrix
(Theorem~\ref{thm1}(3)):
\[
B = M^* \wedge (M^*)^T.
\]

\item Determine the SCC labels $S_{TAG}$:
\begin{enumerate}
\item[(7-1)] Initialize:
$T_{SCC}=1$;
$V_{Undo}=\{1,\dots,n\}$;
\item[(7-2)] While $V_{Undo}\neq\emptyset$ Do
$\{$
\[
i=\min(V_{Undo}),\qquad
V_{SCC}=\{\,j: B(i,j)=1\,\};
\]
and assign
\[
S_{TAG}(v)\Leftarrow T_{SCC} \quad \forall v\in V_{SCC};
\]
update
\[
V_{Undo}\Leftarrow V_{Undo}\setminus V_{SCC},\quad
T_{SCC}\Leftarrow T_{SCC}+1.
\]
$\}$
\end{enumerate}

\item Construct the condensation graph
(steps 8--10 may be implemented as a subprocedure):
\begin{enumerate}
\item[(8-1)] Take the order of the condensation graph as
\[
k=\max_i S_{TAG}(i).
\]
\item[(8-2)] If $k=1$, then directly set
\[
G_{TAG}(i)=1, \quad L_{TAG}(i)=1,\quad I_{TAG}(i)=0,\quad i=1,\dots,n,
\]
and jump to Step 12; otherwise continue.
\item[(8-3)] Initialize condensation-graph-related matrices:
$M_C^{k\times k}=0$, $R_C^{k\times n}=0$.
\item[(8-4)] Construct the condensation compression matrix $R_C$
(Definition~\ref{defn2}(3)):
\[
R_C\bigl(S_{TAG}(i),\,Index(i)\bigr)\Leftarrow 1,\quad i=1,\dots,n.
\]
\item[(8-5)] Generate the condensation-graph adjacency matrix
(Theorem~\ref{thm4}(2)):
\[
M_C = (R_C \odot M \odot R_C^T)\wedge \neg I_k.
\]
\item[(8-6)] Write into the output cache:
\[
C_M(i,j)\Leftarrow M_C(i,j),\quad i,j=1,\dots,k.
\]
\end{enumerate}

\item Process the SCC layer labels $L_{TAG}$
(Theorem~\ref{thm4}(1)):
\begin{enumerate}
\item[(9-1)] Initialize:
$L_{SCC}=1$;
$S_{Comp}=\emptyset$;
\item[(9-2)] Define the mask function $\text{Mask}(S)$:
mask out the rows and columns corresponding to the set $S$,
\[
\text{Mask}(S)(i,j)=
\begin{cases}
0, & i\in S\ \vee\ j\in S,\\
1, & \text{otherwise};
\end{cases}
\]
\item[(9-3)] While $\{1,\dots,k\}\setminus S_{Comp}\neq\emptyset$ Do
$\{$
\[
Y = (M_C \wedge \text{Mask}(S_{Comp}))\odot \mathbf{1};
\quad \mathbf{1}=(1,\dots,1)^T;
\]
\[
S_L=\{\,i\in \{1,\dots,k\}\setminus S_{Comp}:\ Y(i)=0\,\};
\]
assign
\[
L_{TAG}(j)\Leftarrow L_{SCC},
\quad \forall j\in\{1,\dots,n\}\ \text{s.t.}\ S_{TAG}(j)\in S_L ;
\]
update
\[
S_{Comp}\Leftarrow S_{Comp}\cup S_L;\quad
L_{SCC}\Leftarrow L_{SCC}+1
\]
$\}.$
\end{enumerate}

\item Process the weakly connected component labels $G_{TAG}$
(Theorem~\ref{thm1}(4)):
\begin{enumerate}
\item[(10-1)] Construct the underlying graph of the condensation graph
(Definition~\ref{defn1}(3)):
\[
B_C = M_C \vee M_C^T.
\]
\item[(10-2)] Compute $B_C^*$
(Kleene--Valiant):
$$B_C^* = B_C \vee I ;$$
For $t=1$ to $\left\lceil \log_2 (k-1)\right\rceil$ Do
$$B_C^* \Leftarrow B_C^* \odot B_C^* .$$

\item[(10-3)] For an undirected graph,
$B_C^*$ is exactly the mutual-reachability matrix;
subsequently,
$B_C^*(i,j)=1$ is used to determine
whether two SCCs belong to the same weakly connected subgraph.

\item[(10-4)] Partition the SCCs according to $B_C^*$
and write the result back into the node table:

Initialize
$G_{SCC}=1$;
$S_{Undo}=\{1,\dots,k\}$;

While $S_{Undo}\neq\emptyset$ Do
$\{$
\[
i=\min(S_{Undo});\qquad
S_G=\{\,j:\ B_C^*(i,j)=1\,\};
\]
assign
\[
G_{TAG}(v)\Leftarrow G_{SCC}
\quad \forall v\in\{1,\dots,n\}\ \text{s.t.}\ S_{TAG}(v)\in S_G ;
\]
update
\[
S_{Undo}\Leftarrow S_{Undo}\setminus S_G; \quad
G_{SCC}\Leftarrow G_{SCC}+1
\]
$\}.$
\end{enumerate}

\item Structural isolated-point indicator $I_{TAG}$:

For each weakly connected component label $g$,
count the number of nodes it contains, denoted $N_g$.
If $N_g=1$,
set the unique node to $I_{TAG}=1$;
otherwise set $I_{TAG}=0$.

\item Data organization and generation of new indices:

Sort
$V\langle Index,S_{TAG},G_{TAG},L_{TAG},I_{TAG},NewIndex\rangle^n$
in ascending order of
$G_{TAG}$, $I_{TAG}$, $L_{TAG}$, and $S_{TAG}$;
after sorting,
assign sequentially
\[
NewIndex(i)=i,\qquad i=1,\dots,n.
\]

\item Output:
\begin{enumerate}
\item[(13-1)] node attribute table
$V\langle Index,S_{TAG},G_{TAG},L_{TAG},I_{TAG},NewIndex\rangle^n$;
\item[(13-2)] condensation-graph adjacency matrix:
output the upper-left $k\times k$ block of $C_M$;
\item[(13-3)] the order $k$ of the condensation graph.
\end{enumerate}

\end{enumerate}

\subsection{Algorithm Example}
\begin{figure}[H]
  \centering
  \includegraphics[width=0.7\textwidth]{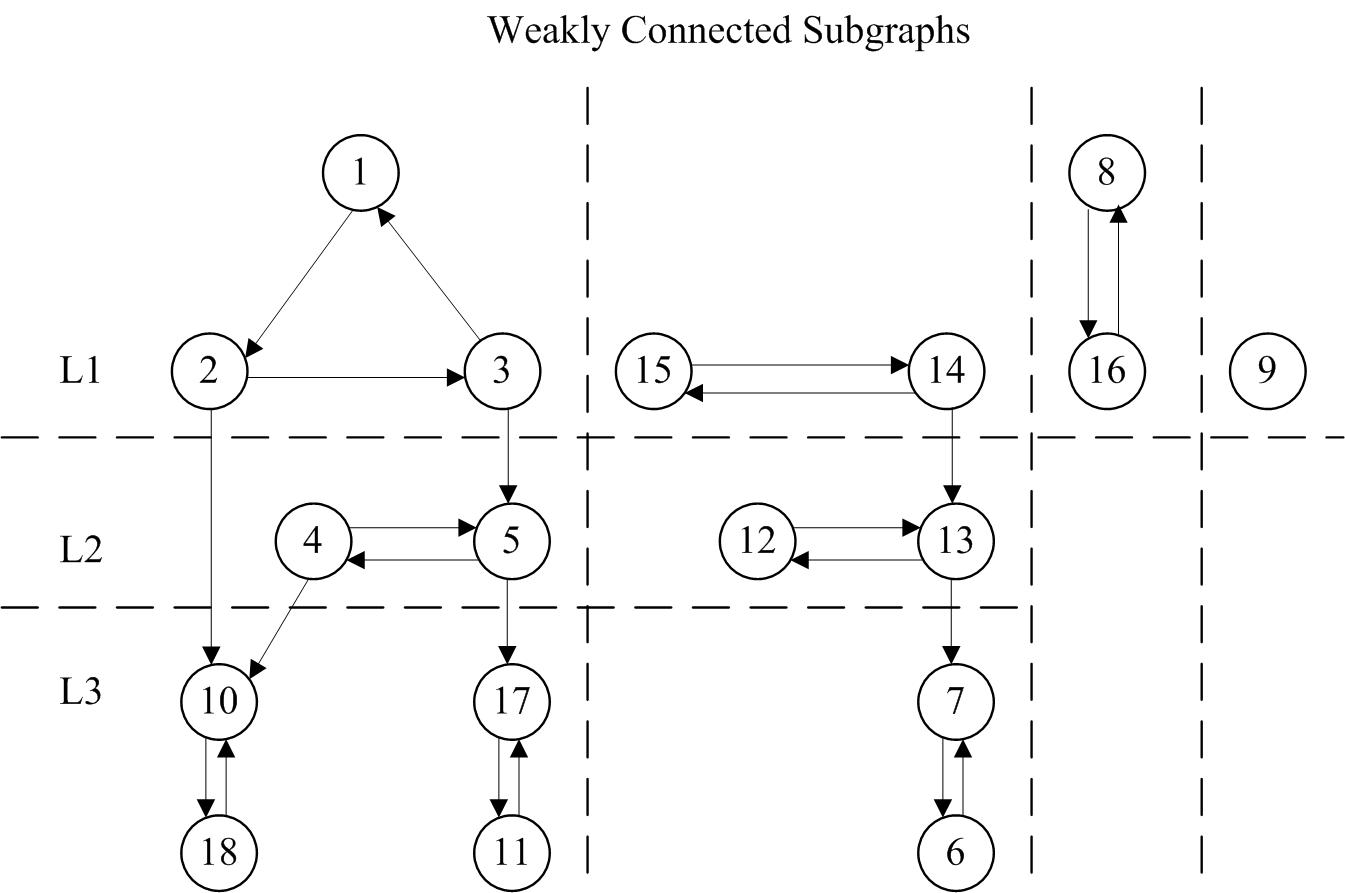}
  \caption{Algorithm verification example: original graph, layer structure, and weakly connected subgraph partition}
  \label{fig:examp0}
\end{figure}

Figure~\ref{fig:examp0} is the original graph of the verification example.
The numbers inside the circles are the original node indices.
Table~\ref{tab:examp0} gives its adjacency matrix.
Under the original indexing,
the nonzero entries of the adjacency matrix $M$
appear globally scattered and irregular.
Although the complete topological information is already contained in it,
one cannot directly observe a clear block structure
corresponding to strongly connected components,
hierarchical levels,
or weakly connected subgraphs.
\begin{table}[H]
\centering
\tiny\ttfamily
\setlength\tabcolsep{2pt}
\renewcommand{\arraystretch}{0.8}
\resizebox{0.7\textwidth}{!}{%
\begin{tabular}{r|cccccccccccccccccc}
 & 1 & 2 & 3 & 4 & 5 & 6 & 7 & 8 & 9 & 10 & 11 & 12 & 13 & 14 & 15 & 16 & 17 & 18 \\ \hline
1 &   &   & 1 &   &   &   &   &   &   &   &   &   &   &   &   &   &   &   \\
2 & 1 &   &   &   &   &   &   &   &   &   &   &   &   &   &   &   &   &   \\
3 &   & 1 &   &   &   &   &   &   &   &   &   &   &   &   &   &   &   &   \\
4 &   &   &   &   & 1 &   &   &   &   &   &   &   &   &   &   &   &   &   \\
5 &   &   & 1 & 1 &   &   &   &   &   &   &   &   &   &   &   &   &   &   \\
6 &   &   &   &   &   &   & 1 &   &   &   &   &   &   &   &   &   &   &   \\
7 &   &   &   &   &   & 1 &   &   &   &   &   &   & 1 &   &   &   &   &   \\
8 &   &   &   &   &   &   &   &   &   &   &   &   &   &   &   & 1 &   &   \\
9 &   &   &   &   &   &   &   &   &   &   &   &   &   &   &   &   &   &   \\
10 &   & 1 &   & 1 &   &   &   &   &   &   &   &   &   &   &   &   &   & 1 \\
11 &   &   &   &   &   &   &   &   &   &   &   &   &   &   &   &   & 1 &   \\
12 &   &   &   &   &   &   &   &   &   &   &   &   & 1 &   &   &   &   &   \\
13 &   &   &   &   &   &   &   &   &   &   &   & 1 &   & 1 &   &   &   &   \\
14 &   &   &   &   &   &   &   &   &   &   &   &   &   &   & 1 &   &   &   \\
15 &   &   &   &   &   &   &   &   &   &   &   &   &   & 1 &   &   &   &   \\
16 &   &   &   &   &   &   &   & 1 &   &   &   &   &   &   &   &   &   &   \\
17 &   &   &   &   & 1 &   &   &   &   &   & 1 &   &   &   &   &   &   &   \\
18 &   &   &   &   &   &   &   &   &   & 1 &   &   &   &   &   &   &   &   \\
\end{tabular}}
\caption{Adjacency matrix $M$ (original node indices)}
\label{tab:examp0}
\end{table}

Table~\ref{tab:finalV} gives the node attribute table
generated after the algorithm terminates.
Here,
$S_{TAG}$ denotes the index of the strongly connected component to which a node belongs,
$L_{TAG}$ denotes the layer index of that SCC in the condensation graph,
$G_{TAG}$ denotes the index of the weakly connected subgraph to which the node belongs,
$I_{TAG}$ is the structural isolated-point indicator,
and $NewIndex$ is the new node index generated according to the sorting rule
$(G_{TAG}, I_{TAG}, L_{TAG}, S_{TAG})$.
\begin{table}[H]
\centering
\small
\setlength\tabcolsep{4pt}
\renewcommand{\arraystretch}{1}
\begin{tabular}{r|rrrrrr}
i & $Index$ & $S_{TAG}$ & $G_{TAG}$ & $L_{TAG}$ & $I_{TAG}$ & $NewIndex$ \\ \hline
1  & 1  & 1 & 1 & 1 & 0  & 1  \\
2  & 2  & 1 & 1 & 1 & 0  & 2  \\
3  & 3  & 1 & 1 & 1 & 0  & 3  \\
4  & 4  & 2 & 1 & 2 & 0  & 4  \\
5  & 5  & 2 & 1 & 2 & 0  & 5  \\ 
6  & 10 & 6 & 1 & 3 & 0  & 6  \\
7  & 18 & 6 & 1 & 3 & 0  & 7  \\
8  & 11 & 7 & 1 & 3 & 0  & 8  \\
9  & 17 & 7 & 1 & 3 & 0  & 9  \\ 
10 & 14 & 9 & 2 & 1 & 0  & 10 \\
11 & 15 & 9 & 2 & 1 & 0  & 11 \\ 
12 & 12 & 8 & 2 & 2 & 0  & 12 \\
13 & 13 & 8 & 2 & 2 & 0  & 13 \\ 
14 & 6  & 3 & 2 & 3 & 0  & 14 \\
15 & 7  & 3 & 2 & 3 & 0  & 15 \\ 
16 & 8  & 4 & 3 & 1 & 0  & 16 \\
17 & 16 & 4 & 3 & 1 & 0  & 17 \\ 
18 & 9  & 5 & 4 & 1 & 1  & 18 \\
\end{tabular}
\caption{Final computational result}
\label{tab:finalV}
\end{table}

\begin{figure}[H]
  \centering
  \includegraphics[width=0.7\textwidth]{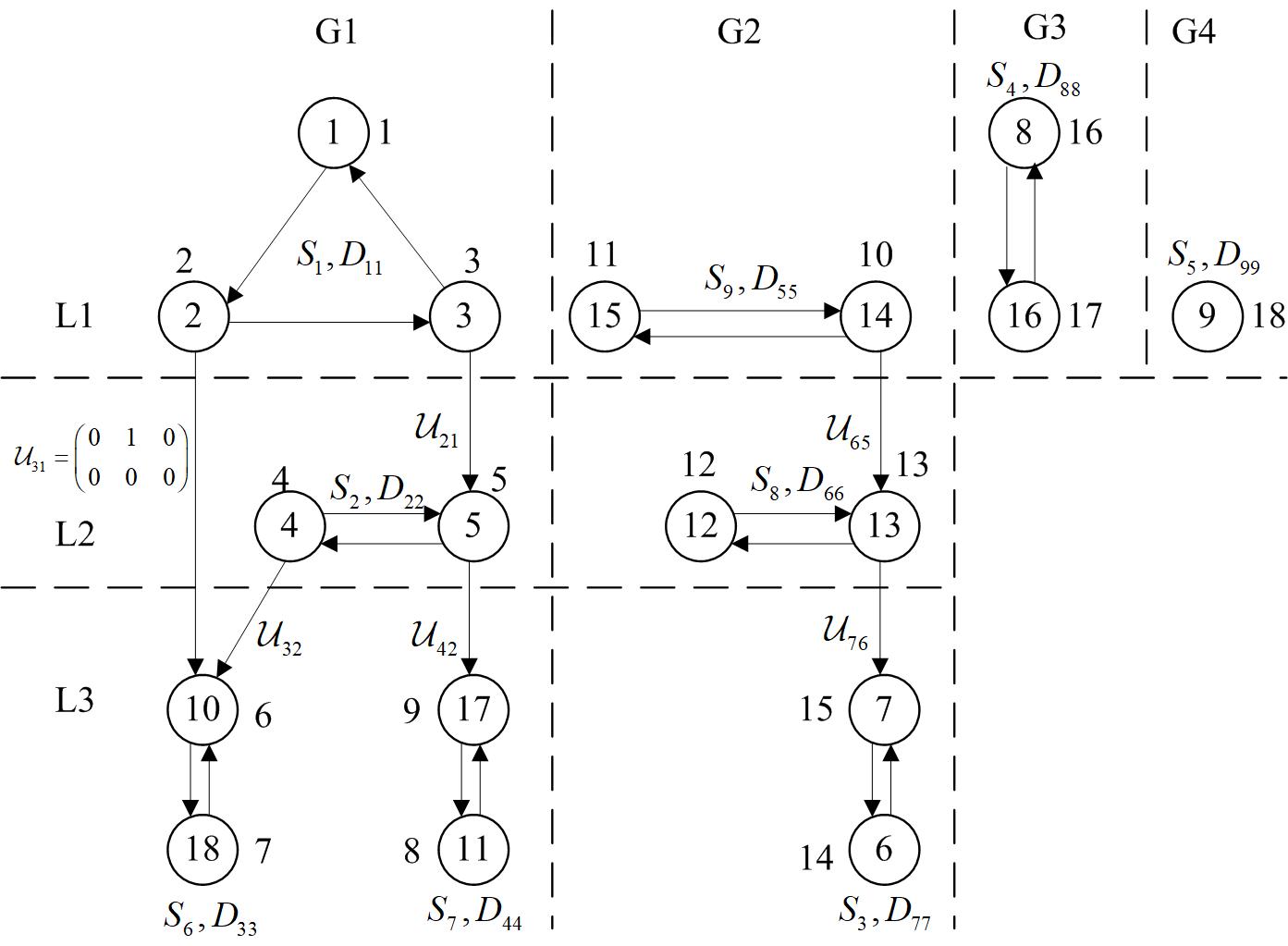}
  \caption{Illustration of the structural partition after permutation}
  \label{fig:example1}
\end{figure}

Figure~\ref{fig:example1} is a graphical illustration
of the result in Table~\ref{tab:finalV}.
The numbers outside the node circles are the new node indices obtained by computation.
Inside each circle,
the subscript $i$ of $S_i$ is the SCC index obtained by computation,
and ${\cal D}_{ii}$ denotes the position index of the diagonal block
after sorting by the new indices.
The leftmost column gives the SCC layer indices obtained by computation,
and the top row gives the weakly connected subgraph indices.
Some edges are labeled by ${\cal U}_{ij}$,
which denotes the communication block from SCC $S_i$ to $S_j$.
From the figure we can observe:

(1) The new node indices are partitioned continuously in the graph:
$G1:1$--$9$, $G2:10$--$15$, $G3:16$--$17$, $G4:18$
(this guarantees that the matrix can be permuted into a block-diagonal structure);

(2) The new node indices of each SCC are consecutive
(i.e., the nodes within the same SCC are adjacent in order),
and the node indices of each SCC exhibit the following feature:
the higher the SCC lies in the hierarchy,
the smaller its new node indices are
(this guarantees that the diagonal blocks can be further permuted
into a lower block-triangular structure);

(3) The unique isolated point in the figure has the final new index $18$,
its corresponding weakly connected subgraph is also the last subgraph $G4$,
and therefore it corresponds to the last diagonal block ${\cal D}_{99}$.
\begin{figure}[H]
  \centering
  \includegraphics[width=0.42\textwidth]{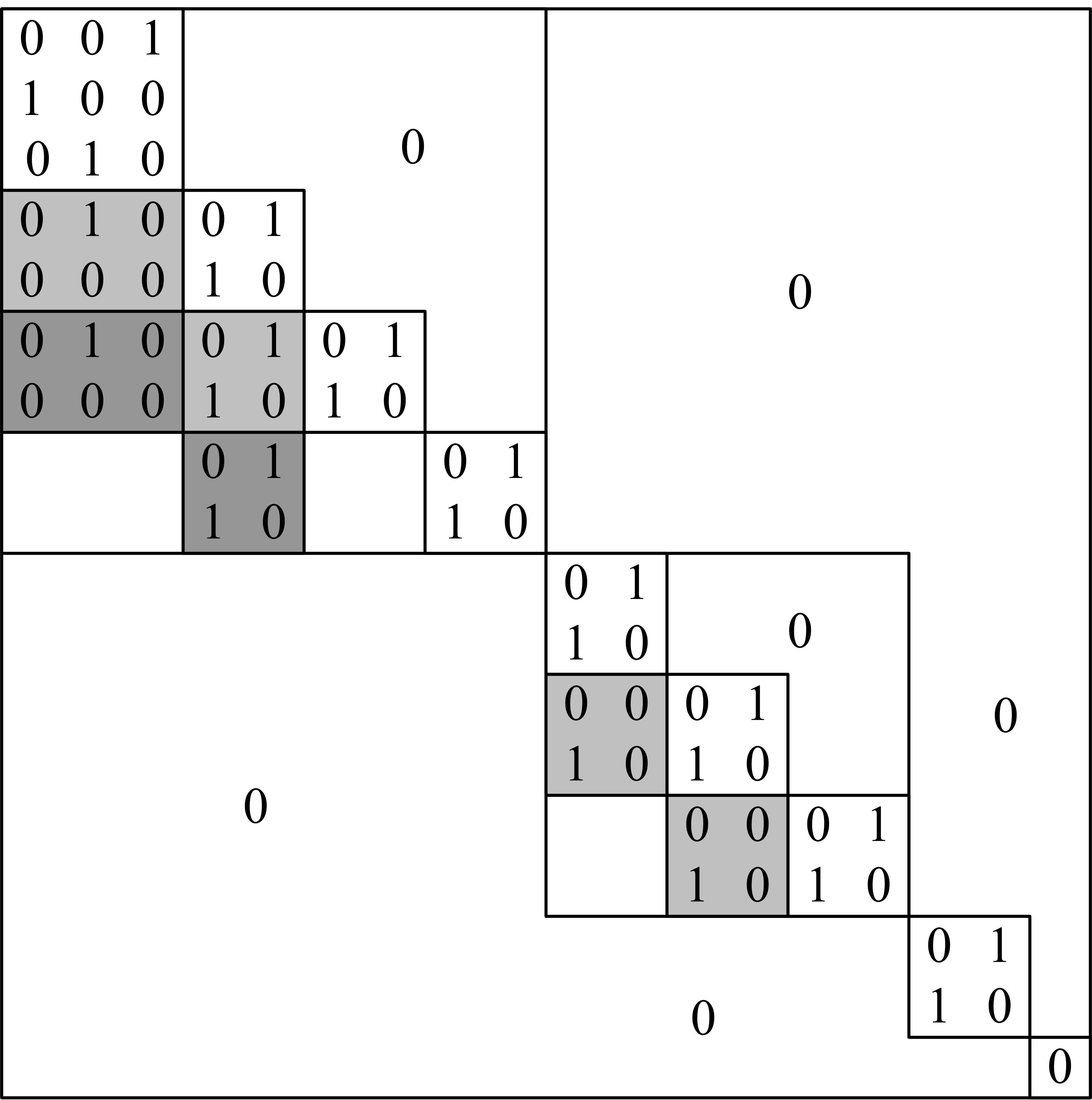}
  \caption{Adjacency matrix permuted by the new node indices}
  \label{fig:example}
\end{figure}

Figure~\ref{fig:example} shows the result
of applying the permutation
\[
P =  < NewIndex ,Index  >
\]
to the adjacency matrix of the original graph
(Table~\ref{tab:examp0}).
Its outer structure is block diagonal,
while inside each diagonal block there is a triangular block structure.
Two degenerate cases appear in the figure:
the bottom-right diagonal block corresponds to the case
where a weakly connected subgraph degenerates into an isolated point
($G5$ degenerates into a singleton),
and the second diagonal block from the bottom-right
corresponds to the case
where a weakly connected subgraph ($G4$)
degenerates into a single strongly connected component.

%
%
%
%
%
%

\FloatBarrier


\begin{thebibliography}{99}

\bibitem{lecun2015deep}
Y.~LeCun, Y.~Bengio, and G.~Hinton,
``Deep learning,''
\emph{Nature},
vol.~521, no.~7553, pp.~436--444, 2015.

\bibitem{brown2020language}
T.~B. Brown \emph{et al.},
``Language Models are Few-Shot Learners,''
\emph{Advances in Neural Information Processing Systems},
vol.~33, pp.~1877--1901, 2020.

\bibitem{wei2022chain}
J.~Wei, X.~Wang, D.~Schuurmans, M.~Bosma, F.~Xia, E.~Chi, Q.~Le, and D.~Zhou,
``Chain-of-Thought Prompting Elicits Reasoning in Large Language Models,''
in \emph{Advances in Neural Information Processing Systems (NeurIPS)},
vol.~35, 2022.

\bibitem{bommasani2021foundation}
R.~Bommasani et~al.,
``On the Opportunities and Risks of Foundation Models,''
\emph{arXiv preprint arXiv:2108.07258}, 2021.

\bibitem{srivastava2023beyond}
A.~Srivastava et al.,
``Beyond the Imitation Game: Quantifying and Extrapolating the Capabilities of Language Models,''
\emph{Transactions on Machine Learning Research (TMLR)}, 2023.

\bibitem{vaswani2017attention}
A.~Vaswani \emph{et al.},
``Attention Is All You Need,''
\emph{Advances in Neural Information Processing Systems},
vol.~30, 2017.

\bibitem{cheng2018model}
Y.~Cheng, D.~Wang, P.~Zhou, and T.~Zhang,
``Model compression and acceleration for deep neural networks:
The principles, progress, and challenges,''
\emph{IEEE Signal Processing Magazine},
vol.~35, no.~1, pp.~126--136, 2018.

\bibitem{narayanan2021efficient}
D.~Narayanan, S.~H. Hashemi, P.~Patel, et al.,
``Efficient large-scale language model training on gpu clusters using megatron-lm,''
in \emph{Proceedings of the 14th USENIX Symposium on Operating Systems Design and Implementation (OSDI)},
2021.

\bibitem{dao2022flashattention}
T.~Dao, D.~Fu, S.~Ermon, A.~R{\'e}, and C.~R{\'e},
``FlashAttention: Fast and Memory-Efficient Exact Attention with IO-Awareness,''
in \emph{Advances in Neural Information Processing Systems (NeurIPS)},
2022.

\bibitem{kaplan2020scaling}
J.~Kaplan, S.~McCandlish, T.~Henighan, T.~B. Brown, B.~Chess,
R.~Child, S.~Gray, A.~Radford, J.~Wu, and D.~Amodei,
``Scaling Laws for Neural Language Models,''
\emph{arXiv preprint arXiv:2001.08361}, 2020.

\bibitem{strubell2019energy}
E.~Strubell, A.~Ganesh, and A.~McCallum,
``Energy and Policy Considerations for Deep Learning in NLP,''
in \emph{Proceedings of ACL}, 2019.

\bibitem{rae2022scaling}
J.~Rae et al.,
``Scaling Language Models: Methods, Analysis \& Insights from Training Gopher,''
\emph{arXiv:2112.11446}, 2022.

\bibitem{chowdhery2022palm}
A.~Chowdhery et al.,
``PaLM: Scaling Language Modeling with Pathways,''
in \emph{Proceedings of the 39th International Conference on Machine Learning (ICML)},
2022.

\bibitem{zhang2017rethinking}
Zhang, Chiyuan and Bengio, Samy and Hardt, Moritz and Recht, Benjamin and Vinyals, Oriol,
``Understanding deep learning requires rethinking generalization,''
\emph{International Conference on Learning Representations}, 2017.

\bibitem{neyshabur2017exploring}
B.~Neyshabur, S.~Bhojanapalli, D.~McAllester, and N.~Srebro,
``Exploring generalization in deep learning,''
\emph{Advances in Neural Information Processing Systems}, vol.~30, 2017.

\bibitem{gunasekar2018implicit}
S.~Gunasekar, J.~D. Lee, D.~Soudry, and N.~Srebro,
``Implicit regularization in matrix factorization,''
\emph{Advances in Neural Information Processing Systems}, vol.~30, 2017.

\bibitem{chizat2020lazy}
L.~Chizat, E.~Oyallon, and F.~Bach,
``On Lazy Training in Differentiable Programming,''
in \emph{Advances in Neural Information Processing Systems (NeurIPS)},
2019.

\bibitem{yang2023feature}
G.~Yang and E.~Hu,
``Feature Learning in Infinite-Width Neural Networks,''
in \emph{arXiv e-prints},
2020.

\bibitem{olah2018building}
Olah, Chris and Satyanarayan, Arvind and Johnson, Ian and Carter, Shan and Schubert, Ludwig and Ye,
``The Building Blocks of Interpretability,''
\emph{Distill}, 2018.

\bibitem{olah2020zoom}
Olah, Chris and Cammarata, Nick and Schubert, Ludwig and Goh, Gabriel and Petrov, Michael and Carter, Shan,
``Zoom in: An introduction to circuits,''
\emph{Distill}, 2020.

\bibitem{elhage2021circuits}
Elhage, Nelson and Nanda, Neel and Olsson, Catherine and Henighan, Tom and Joseph, Nicholas and Mann, Ben and Askell, Amanda and Bai, Yuntao and Chen, Anna and Conerly, Tom and others, 
``A Mathematical Framework for Transformer Circuits,''
\emph{Transformer Circuits Thread}, 2021, vol 1.

\bibitem{zoph2016neural}
B.~Zoph and Q.~V. Le,
``Neural architecture search with reinforcement learning,''
arXiv:1611.01578, 2016.

\bibitem{elsken2019neural}
T.~Elsken, J.~H. Metzen, and F.~Hutter,
``Neural architecture search: A survey,''
\emph{Journal of Machine Learning Research},
vol.~20, no.~55, pp.~1--21, 2019.

\bibitem{wen2016structured}
W.~Wen, C.~Wu, Y.~Wang, Y.~Chen, and H.~Li,
``Learning Structured Sparsity in Deep Neural Networks,''
in \emph{Advances in Neural Information Processing Systems (NeurIPS)},
2016.


\bibitem{han2015learning}
S.~Han, J.~Pool, J.~Tran, and W.~Dally,
``Learning both weights and connections for efficient neural networks,''
\emph{Advances in Neural Information Processing Systems}, vol.~28, 2015.

\bibitem{frankle2019lottery}
J.~Frankle and M.~Carbin,
``The Lottery Ticket Hypothesis: Finding Sparse, Trainable Neural Networks,''
in \emph{International Conference on Learning Representations (ICLR)},
2019.

\bibitem{hoefler2021sparsity}
T.~Hoefler, D.~Alistarh, T.~Ben-Nun, A.~Dryden, and P.~Peste,
``Sparsity in Deep Learning: Pruning and Growth for Efficient Inference and Training,''
\emph{Journal of Machine Learning Research},
vol.~22, no.~241, 2021.

\bibitem{liu2019rethinking}
Z.~Liu, M.~Sun, T.~Zhou, G.~Huang, and T.~Darrell,
``Rethinking the Value of Network Pruning,''
in \emph{International Conference on Learning Representations (ICLR)},
2019.



\bibitem{liu2023sparse}
Frantar, Elias and Alistarh, Dan,
``SparseGPT: Massive Language Models Can Be Accurately Pruned in One-Shot,''
in \emph{Proceedings of the 40th International Conference on Machine Learning (ICML)},
2023.







\bibitem{michel2019sixteen}
P.~Michel, O.~Levy, and G.~Neubig,
``Are Sixteen Heads Really Better than One?''
in \emph{Advances in Neural Information Processing Systems (NeurIPS)},
2019.

\bibitem{voita2019analyzing}
E.~Voita, D.~Talbot, F.~Moiseev, R.~Sennrich, and I.~Titov,
``Analyzing Multi-Head Self-Attention: Specialized Heads Do the Heavy Lifting, the Rest Can Be Pruned,''
in \emph{Proceedings of the 57th Annual Meeting of the Association for Computational Linguistics (ACL)},
pp.~5797--5808, 2019.

\bibitem{vig2019analyzing}
J.~Vig,
``Analyzing the Structure of Attention in a Transformer Language Model,''
in \emph{Proceedings of the Workshop on BlackboxNLP: Analyzing and Interpreting Neural Networks for NLP},
pp.~63--76, 2019.

\bibitem{geva2021transformer}
M.~Geva, R.~Schuster, J.~Berant, and O.~Levy,
``Transformer Feed-Forward Layers Are Key-Value Memories,''
in \emph{Proceedings of the 2021 Conference on Empirical Methods in Natural Language Processing (EMNLP)},
pp.~5484--5495, 2021.

\bibitem{dalvi2020analyzing}
F.~Dalvi, N.~Durrani, H.~Sajjad, and Y.~Belinkov,
``Analyzing Redundancy in Pretrained Transformer Models,''
in \emph{Proceedings of the 2020 Conference on Empirical Methods in Natural Language Processing (EMNLP)},
pp.~442--458, 2020.

\bibitem{goodfellow2016deep}
I.~Goodfellow, Y.~Bengio, and A.~Courville,
\emph{Deep Learning},
MIT Press, 2016.

\bibitem{rumelhart1986learning}
D.~E. Rumelhart, G.~E. Hinton, and R.~J. Williams,
``Learning representations by back-propagating errors,''
\emph{Nature},
vol.~323, pp.~533--536, 1986.

\bibitem{kleene1956representation}
S.~C. Kleene,
``Representation of events in nerve nets and finite automata,''
in \emph{Automata Studies},
C.~E. Shannon and J.~McCarthy, Eds.,
Princeton University Press,
pp.~3--41, 1956.





\bibitem{warshall1962boolean}
S.~Warshall,
``A theorem on Boolean matrices,''
\emph{Journal of the ACM},
vol.~9, no.~1,
pp.~11--12,
1962.





\bibitem{valiant1975general}
L.~G. Valiant,
``General context-free recognition in less than cubic time,''
\emph{Journal of Computer and System Sciences},
vol.~10, no.~2,
pp.~308--315,
1975.




\bibitem{ZweigKaufmann2011}
K.~A.~Zweig and M.~Kaufmann,
\newblock A systematic approach to the one-mode projection of bipartite graphs,
\newblock {\em Social Network Analysis and Mining}, 1(3):187--218, 2011.





\bibitem{Sims1971Computation}
C.~C.~Sims,
\newblock Computation with permutation groups,
\newblock in \emph{Proceedings of the Second Symposium on Symbolic and Algebraic Manipulation},
\newblock 1971, pp.~23--28.

\bibitem{Seress2003Permutation}
Á.~Seress,
\newblock \emph{Permutation Group Algorithms},
\newblock Cambridge Tracts in Mathematics, Vol.~152,
\newblock Cambridge University Press, 2003.






\bibitem{kim1982boolean}
K.~H. Kim and F.~W. Roush,
\emph{Boolean Matrix Theory and Applications},
Marcel Dekker, 1982.





\bibitem{bondy2008graph}
J.~A. Bondy and U.~S.~R. Murty,
\emph{Graph theory with applications},
London: Macmillan, 1976.



\bibitem{hopcroft2001automata}
J.~E. Hopcroft, R.~Motwani, and J.~D. Ullman,
\emph{Introduction to Automata Theory, Languages, and Computation},
Acm Sigact News, 2001, 32(1): 60-65.





\bibitem{cormen2009clrs}
T.~H. Cormen, C.~E. Leiserson, R.~L. Rivest, and C.~Stein,
\emph{Introduction to Algorithms},
MIT Press, 2022.


\bibitem{armstrong1988groups}
M.~A. Armstrong,
\emph{Groups and Symmetry},
Springer Science and Business Media, 1997.




\bibitem{horn2013matrix}
R.~A. Horn and C.~R. Johnson,
\emph{Matrix Analysis},
2nd ed.,
Cambridge University Press, 2012.





%
%
%
%
%
%
%
%
%
%
%
%
%
%
%
%
%
%
%
%
%
%
%
%
%
%
%
%
%
%
%
%
%
%
%
%
%
%
%
%
%
%
%
%
%

%
%
%
%
%
%
%
%
%
%
%
%
%
%
%
%
%
%
%
%
%
%
%
%
%
%
%
%
%
%
%
%
%
%
%
%
%
%
%

\end{thebibliography}
\end{document}